# Consistency Techniques for Flow-Based Projection-Safe Global Cost Functions in Weighted Constraint Satisfaction

**J.H.M. Lee**                                                    JLEE@CSE.CUHK.EDU.HK
**K.L. Leung**                                                 KLLEUNG@CSE.CUHK.EDU.HK
*Department of Computer Science and Engineering*
*The Chinese University of Hong Kong*
*Shatin, N.T., Hong Kong*

## Abstract

Many combinatorial problems deal with preferences and violations, the goal of which is to find solutions with the minimum cost. Weighted constraint satisfaction is a framework for modeling such problems, which consists of a set of cost functions to measure the degree of violation or preferences of different combinations of variable assignments. Typical solution methods for weighted constraint satisfaction problems (WCSPs) are based on branch-and-bound search, which are made practical through the use of powerful consistency techniques such as AC*, FDAC*, EDAC* to deduce hidden cost information and value pruning during search. These techniques, however, are designed to be efficient only on binary and ternary cost functions which are represented in table form. In tackling many real-life problems, high arity (or global) cost functions are required. We investigate efficient representation scheme and algorithms to bring the benefits of the consistency techniques to also high arity cost functions, which are often derived from hard global constraints from classical constraint satisfaction.

The literature suggests some global cost functions can be represented as flow networks, and the minimum cost flow algorithm can be used to compute the minimum costs of such networks in polynomial time. We show that naive adoption of this flow-based algorithmic method for global cost functions can result in a stronger form of $\varnothing$-inverse consistency. We further show how the method can be modified to handle cost projections and extensions to maintain generalized versions of AC* and FDAC* for cost functions with more than two variables. Similar generalization for the stronger EDAC* is less straightforward. We reveal the oscillation problem when enforcing EDAC* on cost functions sharing more than one variable. To avoid oscillation, we propose a weak version of EDAC* and generalize it to weak EDGAC* for non-binary cost functions. Using various benchmarks involving the soft variants of hard global constraints ALLDIFFERENT, GCC, SAME, and REGULAR, empirical results demonstrate that our proposal gives improvements of up to an order of magnitude when compared with the traditional constraint optimization approach, both in terms of time and pruning.

## 1. Introduction

Constraint satisfaction problems (CSPs) occur in all walks of industrial applications and computer science, such as scheduling, bin packing, transport routing, type checking, diagram layout, just to name a few. Constraints in CSPs are functions returning true or false. These constraints are hard in the sense that they must be satisfied. In over-constrained and optimization scenarios, hard constraints have to be relaxed or softened. The weighted constraint satisfaction framework adopt soft constraints as *cost functions* returning a non-negative integer with an upper bound $\top$. Solution techniques for solving weighted constraint satisfaction problems (WCSPs) are made practi-cal by enforcing various consistency notions during branch-and-bound search, such as NC*, AC*, FDAC* (Larrosa & Schiex, 2004, 2003) and EDAC* (de Givry, Heras, Zytnicki, & Larrosa, 2005). These enforcement techniques, however, are designed to be efficient only on binary and ternary cost functions which are represented in table form. On the other hand, many real-life problems can be modelled naturally by global cost functions of high arities. *We investigate efficient representation scheme and algorithms to bring the benefits of the existing consistency techniques for binary and ternary cost functions to also high arity cost functions, which are often derived from hard global constraints from classical constraint satisfaction.*

In existing WCSP solvers, these high arity cost functions are delayed until they become binary or ternary during search. The size of the tables is also a concern. The lack of efficient handling of high arity global cost functions in WCSP systems greatly restricts the applicability of WCSP techniques to more complex real-life problems. To overcome the difficulty, we incorporate van Hoeve, Pesant, and Rousseau's (2006) flow-based algorithmic method into WCSPs, which amounts to representing global cost functions as flow networks and computing the minimum costs of such networks using the minimum cost flow algorithm. We show that a naive incorporation of global cost functions into WCSPs would result in a strong form of the $\varnothing$-inverse consistency (Zytnicki, Gaspin, & Schiex, 2009), which is still relatively weak in terms of lower bound estimation and pruning. The question is then whether we can *achieve stronger consistencies such as GAC* and FDGAC*, the generalized versions of AC* and FDAC* respectively, for non-binary cost functions efficiently.* Consistency algorithms for (G)AC* and FD(G)AC* involve three main operations: (a) computing the minimum cost of the cost functions when a variable $x$ is fixed with value $v$, (b) projecting the minimum cost of a cost function to the unary cost functions for $x$ at value $v$, and (c) extending unary costs to the related high arity cost functions. These operations allow cost movements among cost functions and shifting of costs to increase the global lower bound of the problem, which implies more opportunities for domain value prunings. Part (a) is readily handled using the minimum cost flow (MCF) algorithm as proposed in van Hoeve et al.'s method. However, parts (b) and (c) modify the cost functions, which can possibly destroy the required flow-based structure of the cost functions required by van Hoeve et al.'s method. To overcome the difficulty, we propose and give sufficient conditions for the flow-based projection-safety property. If a global cost function is flow-based projection-safe, the flow-based property of the cost function is guaranteed to be retained no matter how many times parts (b) and (c) are performed. Thus, the MCF algorithm can be applied throughout the enforcements of GAC* and FDGAC* to increase search efficiency.

A natural next step is to generalize also the stronger consistency EDAC* (de Givry et al., 2005) to EDGAC*, but this turns out to be non-trivial. We identify and analyze an inherent limitation of EDAC* similar to the case of Full AC* (de Givry et al., 2005). ED(G)AC* enforcement will go into oscillation if two cost functions share more than one variable, which is common when a problem involves high arity cost functions. Sanchez, de Givry, and Schiex (2008) did not mention the oscillation problem but their method for enforcing EDAC* for the special case of ternary cost functions would avoid the oscillation problem. In this paper, we give a weak form of EDAC*, which can be generalized to weak EDGAC* for cost functions of *any* arity. Most importantly, weak EDAC* is reduced to EDAC* when no two cost functions share more than one variable. Weak EDGAC* is stronger than FDGAC* and GAC*, but weaker than VAC (Cooper, de Givry, Sanchez, Schiex, Zytnicki, & Werner, 2010). We also give an efficient algorithm to enforce weak EDGAC*.

Based on the theoretical results, we prove that some of the soft variants of ALLDIFFERENT, GCC, SAME, and REGULAR constraints are flow-based projection-safe, and give polynomial time algorithms to enforce GAC*, FDGAC* and also weak EDGAC* on these cost functions. Experiments are carried out on different benchmarks featuring the proposed global cost functions. Empirical results coincide with the theoretical prediction on the relative strengths of the various consistency notions and the complexities of the enforcement algorithms. Our experimental results also confirm that stronger consistencies such as GAC*, FDGAC* and weak EDGAC* are worthwhile and essential in making global cost functions in WCSP practical. In addition, the reified approach (Petit, Régin, & Bessière, 2000) and strong $\varnothing$IC are too weak in estimating useful lower bounds and pruning the search space in branch-and-bound search.

The rest of the paper is organized as follows. Section 2 gives the necessary definitions and background, while Section 3 gives related work. Generalized versions of existing consistency techniques for global cost functions are presented and compared in Section 4. Enforcement algorithms for these consistencies are exponential in general. We introduce the notion of flow-based projection-safety, and describe polynomial time consistency enforcement algorithms for global cost functions enjoying the flow-based projection-safety property. In Section 5, we prove that the softened form of some common hard global constraints are flow-based projection-safe and give experimental results demonstrating the feasibility and efficiency of our proposal both in terms of runtime and search space pruning. Section 6 summarizes our contributions and shed light on possible directions for future research.

## 2. Background

We give the preliminaries on weighted constraint satisfaction problems, global cost functions and network flows.

### 2.1 Weighted Constraint Satisfaction

A weighted constraint satisfaction problem (WCSP) is a special case of valued constraint satisfaction (Schiex, Fargier, & Verfaillie, 1995) with a cost structure $([0, \ldots, \top], \oplus, \leq)$. The structure contains a set of integers from 0 to $\top$ ordered by the standard ordering $\leq$. Addition $\oplus$ is defined by $a \oplus b = min(\top, a + b)$, and subtraction $\ominus$ is defined only for $a \geq b$, $a \ominus b = a - b$ if $a \neq \top$ and $\top \ominus a = \top$ for any $a$. Formally,

**Definition 1** *(Schiex et al., 1995) A* WCSP *is a tuple* $(\mathcal{X}, \mathcal{D}, \mathcal{C}, \top)$, *where:*

- $\mathcal{X}$ *is a set of variables* $\{x_1, x_2, \ldots, x_n\}$ *ordered by their indices;*

- $\mathcal{D}$ *is a set of domains* $D(x_i)$ *for* $x_i \in \mathcal{X}$, *only one value of which can be assigned to* $x_i$;

- $\mathcal{C}$ *is a set of cost functions* $W_S$ *with different scope* $S = \{x_{s_1}, \ldots, x_{s_n}\} \subseteq \mathcal{X}$ *that maps a tuple* $\ell \in L(S)$, *where* $L(S) = D(x_{s_1}) \times \ldots D(x_{s_n})$, *to* $[0, \ldots, \top]$.

An *assignment* of a set of variables $S \subseteq \mathcal{X}$, written as $\{x_{s_1} \mapsto v_{s_1}, \ldots, x_{s_n} \mapsto v_{s_n}\}$, is to assign each variable $x_{s_i} \in S$ to a value $v_{s_i} \in D(x_{s_i})$. When the context is clear and assuming an ordering by the variable indices, we abuse notations by considering an assignment also a tuple $\ell = (v_{s_1}, \ldots, v_{s_n}) \in \mathcal{L}(S)$, where $\mathcal{L}(S) = D(x_{s_1}) \times D(x_{s_2}) \times \ldots D(x_{s_n})$. The notation $\ell[x_{s_i}]$ denotes the value $v_{s_i}$ assigned to $x_{s_i} \in S$, and $\ell[S']$ denotes the tuple formed by projecting $\ell$ onto $S' \subseteq S$.

Without loss of generality, we assume $\mathcal{C} = \{W_\varnothing\} \cup \{W_i \mid x_i \in \mathcal{X}\} \cup \mathcal{C}^+$. $W_\varnothing$ is a constant nullary cost function. $W_i$ is a unary cost function associated with each $x_i \in \mathcal{X}$. $\mathcal{C}^+$ is a set of cost functions $W_S$ with scope $S$ containing two or more variables. If $W_\varnothing$ and $\{W_i\}$ are not defined, we assume $W_i(v) = 0$ for all $v \in D(x_i)$ and $W_\varnothing = 0$. To simplify the notation, we denote $W_{s_1, s_2, \ldots, s_n}$ for the cost function on variables $\{x_{s_1}, x_{s_2}, \ldots, x_{s_n}\}$ if the context is clear.

**Definition 2** *Given a WCSP $(\mathcal{X}, \mathcal{D}, \mathcal{C}, \top)$. The* cost *of a tuple $\ell \in \mathcal{L}(\mathcal{X})$ is defined as $cost(\ell) = W_\varnothing \oplus \bigoplus_{x_i \in \mathcal{X}} W_i(\ell[x_i]) \oplus \bigoplus_{W_S \in \mathcal{C}^+} W_S(\ell[S])$. A tuple $\ell \in \mathcal{L}(\mathcal{X})$ is* feasible *if $cost(\ell) < \top$, and is a* solution *of a WCSP if $cost(\ell)$ is minimum among all tuples in $\mathcal{L}(\mathcal{X})$.*

WCSPs are usually solved with basic branch-and-bound search augmented with consistency techniques which prune infeasible values from variable domains and push costs into $W_\varnothing$ while preserving the equivalence of the problems, *i.e.* the cost of each tuple $\ell \in \mathcal{L}(\mathcal{X})$ is unchanged. Different consistency notions have been defined such as NC*, AC*, FDAC* (Larrosa & Schiex, 2004, 2003), and EDAC* (de Givry et al., 2005).

**Definition 3** *A variable $x_i$ is* node consistent *(NC*) if each value $v \in D(x_i)$ satisfies $W_i(v) \oplus W_\varnothing < \top$ and there exists a value $v' \in D(x_i)$ such that $W_i(v') = 0$. A WCSP is NC* iff all variables are NC*.*

Procedure `enforceNC*()` in Algorithm 1 enforces NC*, where `unaryProject()` moves unary costs towards $W_\varnothing$ while keeping the solution unchanged, and `pruneVal()` removes infeasible values. The variables $\mathbb{Q}$, $\mathbb{R}$, and $\mathbb{S}$ are global propagation queues used for further consistency enforcements explained in later sections. They are initially empty if not specified.

> **Procedure** `enforceNC*()`
> 1    **foreach** $x_i \in \mathcal{X}$ **do** $unaryProject(x_i)$;
> 2    $pruneVal()$;
>
> **Procedure** `unaryProject($x_i$)`
> 3    $\alpha := \min\{W_i(v) \mid v \in D(x_i)\}$;
> 4    $W_\varnothing := W_\varnothing \oplus \alpha$;
> 5    **foreach** $v \in D(x_i)$ **do** $W_i(v) := W_i(v) \ominus \alpha$;
>
> **Procedure** `pruneVal()`
> 6    **foreach** $x_i \in \mathcal{X}$ **do**
> 7      $flag := \textbf{false}$;
> 8      **foreach** $v \in D(x_i)$ *s.t.* $W_i(v) \oplus W_\varnothing = \top$ **do**
> 9        $D(x_i) := D(x_i) \setminus \{v\}$;
> 10        $flag := \textbf{true}$;
> 11      **if** $flag$ **then**
>        // For further consistency enforcement. Assume initially
>          empty if not specified
> 12        $\mathbb{Q} := \mathbb{Q} \cup \{x_i\}$;
> 13        $\mathbb{S} := \mathbb{S} \cup \{x_i\}$;
> 14        $\mathbb{R} := \mathbb{R} \cup \{x_i\}$;

**Algorithm 1**: Enforce NC*

Based on NC*, AC* and FDAC* have been developed for binary (Larrosa & Schiex, 2004, 2003) and ternary cost functions (Sanchez et al., 2008). Enforcing these consistency notions requires two equivalence preserving transformations besides NC* enforcement, namely *projection and extension* (Cooper & Schiex, 2004).

A *projection*, written as `Project(W_S, W_i, v, α)`, transforms $(W_S, W_i)$ to $(W_S', W_i')$ with respect to a value $v \in D(x_i)$ and a cost $\alpha$, where $\alpha \le \min\{W_S(\ell) \mid \ell[x_i] = v \wedge \ell \in \mathcal{L}(S)\}$, such that:

- $W_i'(u) = \begin{cases} W_i(u) \oplus \alpha & \text{if } u = v, \\ W_i(u) & \text{otherwise.} \end{cases}$

- $W_S'(\ell) = \begin{cases} W_S(\ell) \ominus \alpha & \text{if } \ell[x_i] = v, \\ W_S(\ell) & \text{otherwise.} \end{cases}$

An *extension*, written as `Extend(W_S, W_i, v, α)`, transforms $(W_S, W_i)$ to $(W_S'', W_i'')$ with respect to a value $v \in D(x_i)$ and a cost $\alpha$, where $\alpha \le W_i(v)$, such that:

- $W_i''(u) = \begin{cases} W_i(u) \ominus \alpha & \text{if } u = v, \\ W_i(u) & \text{otherwise.} \end{cases}$

- $W_S''(\ell) = \begin{cases} W_S(\ell) \oplus \alpha & \text{if } \ell[x_i] = v, \\ W_S(\ell) & \text{otherwise.} \end{cases}$

## 2.2 Global Constraints and Global Cost Functions

A *global constraint* is a constraint with special semantics. They are usually with high arity, and thus cannot be propagated efficiently with standard consistency algorithms. With their special semantics, special propagation algorithms can be designed to achieve efficiency.

A *global cost function* is the soft variant of a hard global constraint. The cost of each tuple indicates how much the tuple violates the corresponding global constraint. One global constraint can give rise to different global cost functions using different *violation measures*. A global cost function returns 0 if the tuple satisfies the corresponding global constraint. The notation $\textsc{soft\_GC}^\mu$ denotes the global cost function derived from a global constraint GC using a violation measure $\mu$. For instance, the ALLDIFFERENT constraint has two soft variants.

**Definition 4** *(Petit, Régin, & Bessière, 2001) The cost function* $\textsc{soft\_allDifferent}^{var}$ *returns the minimum number of variable assignments that needed to be changed so that the tuple contains only distinct values; while* $\textsc{soft\_allDifferent}^{dec}$ *returns the number of pairs of variables having the same assigned value.*

## 2.3 Flow Theory

**Definition 5** *A flow network* $G = (V, E, w, c, d)$ *is a connected directed graph* $(V, E)$*, in which each edge* $e \in E$ *has a weight* $w_e$*, a capacity* $c_e$*, and a demand* $d_e \le c_e$*.*

An $(s, t)$-*flow* $f$ *from a source* $s \in V$ *to a sink* $t \in V$ *of a value* $\alpha$ *in* $G$ *is defined as a mapping from* $E$ *to real numbers such that:*

- $\sum_{(s,u) \in E} f_{(s,u)} = \sum_{(u,t) \in E} f_{(u,t)} = \alpha$;

- $\sum_{(u,v) \in E} f_{(u,v)} = \sum_{(v,u) \in E} f_{(v,u)} \; \forall \, v \in V \setminus \{s, t\}$;

- $d_e \le f_e \le c_e \ \forall \, e \in E$.

For simplicity, we call an $(s,t)$-flow as a flow if $s$ and $t$ have been specified.

**Definition 6** *The* cost *of a flow $f$ is defined as $cost(f) = \sum_{e \in E} w_e f_e$. A minimum cost flow problem of a value $\alpha$ is to find the flow whose value is $\alpha$ and cost is minimum.*

If $\alpha$ is not given, it is assumed to be the maximum value among all flows.

To solve minimum cost flow problems, various approaches have been developed. Two of those are the *successive shortest path* and *cycle-cancelling algorithms* (Lawler, 1976). Both algorithms focus on the computation in the residual network of the corresponding flow network.

**Definition 7** *Given a flow $f$ in the network $G = (V, E, w, c, d)$. The* residual network *$G^{res} = (V, E^{res}, w^{res}, c^{res}, d^{res})$ is defined as:*

- $E^{res} = \{(u,v) \in e \mid f_{(u,v)} < c_{(u,v)}\} \cup \{(v,u) \in e \mid f_{(u,v)} > d_{(u,v)}\}$;

- $w_{(u,v)}^{res} = \begin{cases} w_{(u,v)} & \text{,if } f_{(u,v)} < c_{(u,v)} \\ -w_{(u,v)} & \text{,if } f_{(v,u)} > d_{(v,u)} \end{cases}$

- $c_{(u,v)}^{res} = \begin{cases} c_{(u,v)} - f_{(u,v)} & \text{,if } f_{(u,v)} < c_{(u,v)} \\ f_{(u,v)} - d_{(u,v)} & \text{,if } f_{(v,u)} > d_{(v,u)} \end{cases}$

- $d_e^{res} = 0$, for all $e \in E$;

The successive shortest path algorithm successively increases flow values of the edges along the shortest paths between $s$ and $t$ in the residual network until the value of flow reaches $\alpha$ or no more paths can be found. The cycle-cancelling algorithm reduces the cost of the given flow to minimum by removing negative cycles in the induced residual network.

In consistency enforcement with flow, we usually deal with the following problem: consider a $(s,t)$-flow $f$ in a network $G = (V, E, w, c, d)$ with minimum cost, and an edge $\bar{e} \in E$. The problem is to determine whether increasing (or decreasing) $f_{\bar{e}}$ by one unit keeps the flow value unchanged, and compute the minimal cost of the new resultant flow if possible. Again, such a problem can be solved using the residual network $G^{res}$ (Régin, 2002; van Hoeve et al., 2006): we compute the shortest path $P$ from $v'$ to $u'$ in $G^{res}$, where $\bar{e} = (u', v') \in E$. If $P$ exists, the value of the flow is unchanged if $f_{\bar{e}}$ is increased by one unit. The new minimum cost can be computed by the following theorem.

**Theorem 1** *(Régin, 2002; van Hoeve et al., 2006) Suppose $f'$ is the resultant flow by increasing $f_{\bar{e}}$ by one unit. Then the minimum value of $cost(f')$ is $cost(f) + w_{\bar{e}}^{res} + \sum_{e \in P} w_e^{res}$.*

Theorem 1 reduces the problem into finding a shortest path from $v'$ to $u'$, which can be made incremental for consistency enforcement. If we want to reduce a unit flow from an edge, we can apply similar methods to those used in Theorem 1.

## 3. Related Work

Global cost functions can be handled using constraint optimization, which focuses on efficient computation of $\min\{W_S(\ell) \mid \ell \in \mathcal{L}(S)\}$ and enforcing GAC on their hard constraint forms $W_S(\ell) \le z_S$, where $z_S$ is the variable storing costs (Petit et al., 2001). Van Hoeve et al. (2006)

develop a framework for global cost functions representable by flow networks, whose computation is polynomial in the size of networks. Beldiceanu (2000) and Beldiceanu, Carlsson and Petit (2004) further develop a representation scheme for global cost functions using a graph-based approach and an automaton approach. Under their framework, the computation of all global cost functions can be reduced to only considering a fixed set of global cost functions, e.g. the SOFT_REGULAR functions.

On the other hand, to efficiently remove more search space during WCSPs solving, various consistency notions have been developed. Examples are NC* (Larrosa & Schiex, 2004), BAC$^\varnothing$ (Zytnicki et al., 2009), AC* (Larrosa & Schiex, 2004), FDAC* (Larrosa & Schiex, 2003), and EDAC* (de Givry et al., 2005). Stronger consistency notions, namely OSAC and VAC (Cooper et al., 2010), are also defined, but enforcement requires a relaxation of the cost valuation structure $V(\top)$ to rational numbers, and current implementations are efficient only on binary WCSPs. For ternary cost functions, AC, FDAC and EDAC are introduced (Sanchez et al., 2008). Cooper (2005) incorporates the concept of $k$-consistency into WCSPs to form *complete $k$-consistency*. However, the time and space complexities increase exponentially as the problem size increases, making complete $k$-consistency impractical to enforce for general WCSPs.

## 4. Consistency Notions for Global Cost Functions

In this section, we discuss four consistency notions for high-arity cost functions: (1) strong $\varnothing$-inverse consistency (strong $\varnothing$IC), (2) generalized arc consistency (GAC*), (3) full directional generalized arc consistency(FDGAC*), and (4) generalized EDAC*. These consistency notions require exponential time to enforce in general, but flow-based global cost functions (van Hoeve et al., 2006) enjoy polynomial time enforcement.

### 4.1 Strong $\varnothing$-Inverse Consistency

Strong $\varnothing$-inverse consistency is based on $\varnothing$-inverse consistency ($\varnothing$IC) (Zytnicki et al., 2009).

**Definition 8** *(Zytnicki et al., 2009) Given a WCSP $P = (\mathcal{X}, \mathcal{D}, \mathcal{C}, \top)$. A cost function $W_S \in \mathcal{C}$ is $\varnothing$-inverse consistent ($\varnothing$IC) if there exists a tuple $\ell \in \mathcal{L}(S)$ such that $W_S(\ell) = 0$. A WCSP is $\varnothing$IC iff all cost functions are $\varnothing$IC.*

The procedure enforce$\varnothing$IC() in Algorithm 2 enforces $\varnothing$IC. Each cost function $W_S$ is made $\varnothing$IC by lines 3 to 6, which move costs from $W_S$ to $W_\varnothing$ by simple arithmetic operations.

**Function** enforce$\varnothing$IC()
1    $flag :=$ **false**;
2    **foreach** $W_S \in \mathcal{C}$ **do**
3      $\alpha := min\{W_S(\ell) \mid \ell \in \mathcal{L}(S)\}$;
4      $W_\varnothing := W_\varnothing \oplus \alpha$;
5      **foreach** $\ell \in \mathcal{L}(S)$ **do** $W_S(\ell) := W_S(\ell) \ominus \alpha$;
6      **if** $\alpha > 0$ **then** $flag :=$ **true**;
7    **return** $flag$;

**Algorithm 2**: Enforcing $\varnothing$IC on a WCSP

The time complexity of enforce$\varnothing$IC() in Algorithm 2 depends on the time complexities of lines 3 and 5. Line 3 computes the minimum cost and line 5 modifies the cost of each tuple to maintain equivalence. In general, these two operations are exponential in the arity of the cost function. However, the first operation can be reduced to polynomial time for a global cost function. One such example is *flow-based global cost functions* (van Hoeve et al., 2006).

**Definition 9** *(van Hoeve et al., 2006) A global cost function $W_S$ is* flow-based *if $W_S$ can be represented as a flow network $G = (V, E, w, c, d)$ such that*

$$min\{cost(f) \mid f \text{ is the max. } \{s,t\}\text{-flow of } G\} = min\{W_S(\ell) \mid \ell \in \mathcal{L}(S)\},$$

*where $s \in V$ is the fixed source and $t \in V$ is the fixed destination.*

For examples, the cost function SOFT_ALLDIFFERENT$^{dec}(S)$ returns the number of pairs of variables in $S$ that share the same value, and is shown to be flow-based (van Hoeve et al., 2006). An example of its corresponding flow network, where $S = \{x_1, x_2, x_3, x_4\}$, is shown in Figure 1. All edges have a capacity of 1. The numbers on the edges represent the weight of the edges. If an edge has no number, the edge has zero weight. The thick lines show the flow corresponding to the tuple $\ell = (a, c, b, b)$ having a cost of 1.

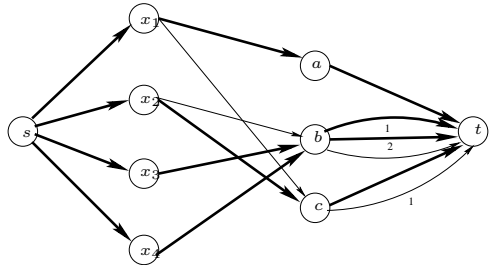

Figure 1: An example flow network for SOFT_ALLDIFFERENT$^{dec}$

With flow-based cost functions, the first operation (computing the minimum cost) can be reduced to time polynomial to the network size for those constraints. The second operation can be reduced to constant time using the $\Delta_S$ data structure suggested by Zytnicki et al. (2009). Instead of deducting the projected value $\alpha$ from each tuple in $W_S$, we simply store the projected value in $\Delta_S$. When we want to know the actual value of $W_S$, we compute $W_S \ominus \Delta_S$.

Enforcing $\varnothing$IC only increases $W_\varnothing$ but does not help reduce domain size. Consider the WCSP in Figure 2. It is $\varnothing$IC, but the value $c \in D(x_1)$ cannot be a part of any feasible tuple. All tuples associated with the assignment $\{x_1 \mapsto c\}$ must have a cost of at least 4: 1 from $W_\varnothing$, 2 from $W_1$, and 1 from $W_{1,2}$. To allow domain reduction, extra conditions are added to $\varnothing$IC to form *strong $\varnothing$IC*.

$$\top = 4, W_\varnothing = 1$$

| $x_1$ | $W_1$ |
|---|---|
| $a$ | 0 |
| $b$ | 2 |
| $c$ | 2 |

| $x_2$ | $W_2$ |
|---|---|
| $a$ | 1 |
| $b$ | 0 |

| $x_1$ | $x_2$ | $W_{1,2}$ |
|---|---|---|
| $a$ | $a$ | 0 |
| $b$ | $a$ | 0 |
| $c$ | $a$ | 1 |

| $x_1$ | $x_2$ | $W_{1,2}$ |
|---|---|---|
| $a$ | $b$ | 0 |
| $b$ | $b$ | 0 |
| $c$ | $b$ | 1 |

Figure 2: A WCSP which is $\varnothing$IC

**Definition 10** *Given a WCSP $P = (\mathcal{X}, \mathcal{D}, \mathcal{C}, \top)$. Consider a non-unary cost function $W_S \in \mathcal{C}^+$ and a variable $x_i \in S$. A tuple $\ell \in \mathcal{L}(S)$ is the $\varnothing$-support of a value $v \in D(x_i)$ with respect to $W_S$ iff $\ell[x_i] = v$ and $W_\varnothing \oplus W_i(v) \oplus W_S(\ell) < \top$. The cost function $W_S$ is strong $\varnothing$IC iff it is $\varnothing$IC, and each value in each variable in $S$ has a $\varnothing$-support with respect to $W_S$. A WCSP is strong $\varnothing$IC if it is $\varnothing$IC and all non-unary cost functions are strong $\varnothing$IC.*

For instance, the WCSP in Figure 2 is not strong $\varnothing$IC. The value $c \in D(x_1)$ does not have a $\varnothing$-support, since $W_\varnothing \oplus W_1(c) \oplus \min\{W_{1,2}(\ell) \mid \ell[x_1] = c \wedge \ell \in \mathcal{L}(\{x_1, x_2\})\} = \top = 4$. Removal of $c \in D(x_1)$ makes it so.

Strong $\varnothing$IC collapses to GAC in classical CSPs when WCSPs collapse to CSPs. Although its definition is similar to BAC$^\varnothing$ (Zytnicki et al., 2009), their strengths are incomparable. BAC$^\varnothing$ gathers cost information from all cost functions on the boundary values, while we only consider the information from one non-unary cost function for all individual values.

The procedure `enforceS∅IC()` in Algorithm 3 enforces strong $\varnothing$IC, based on the `W-AC*3()` Algorithm (Larrosa & Schiex, 2004). The algorithm maintains a propagation queue $\mathbb{Q}$ of variables. Cost functions involving variables in $\mathbb{Q}$ are potentially not strong $\varnothing$IC. At each iteration, an arbitrary variable $x_j$ is removed from $\mathbb{Q}$ by the function pop() in constant time. The algorithm enforces strong $\varnothing$IC for the cost functions involving $x_j$ from lines 4 to 6. The existence of $\varnothing$-support is enforced by `find∅Support()`. If domain reduction occurs (`find∅Support()` returns true), or $W_\varnothing$ increases (`enforce∅IC()` returns true), variables are pushed onto $\mathbb{Q}$ at lines 6 and 7 respectively, indicating that $\varnothing$IC are potentially broken. If the algorithm terminates, i.e. $\mathbb{Q} = \varnothing$, no variables are pushed into $\mathbb{Q}$ at line 6, or $\mathbb{Q}$ is not set to $\mathcal{X}$ at line 7. It implies all variables are strong $\varnothing$IC and the WCSP is $\varnothing$IC. Thus the WCSP is strong $\varnothing$IC after execution.

---

**Procedure** `enforceS∅IC()`

1    $\mathbb{Q} := \mathcal{X}$;
2    **while** $\mathbb{Q} \neq \varnothing$ **do**
3       $x_j := pop\,(\mathbb{Q})$;
4       **foreach** $W_S \in \mathcal{C}^+$ *s.t.* $x_j \in S$ **do**
5          **foreach** $x_i \in S \setminus \{x_j\}$ **do**
6             **if** $find∅Support\,(W_S, x_i)$ **then** $\mathbb{Q} := \mathbb{Q} \cup \{x_i\}$;
7       **if** $enforce∅IC\,()$ **then** $\mathbb{Q} := \mathcal{X}$;

**Function** `find∅Support(`$W_S$`, `$x_i$`)`

8    $flag :=$ **false**;
9    **foreach** $v \in D(x_i)$ **do**
10       $\alpha := min\{W_S(\ell) \mid \ell[x_i] = v\}$;
11       **if** $W_\varnothing \oplus W_i(v) \oplus \alpha = \top$ **then**
12          $D(x_i) := D(x_i) \setminus \{v\}$;
13          $flag :=$ **true**;

14    **return** $flag$;

**Algorithm 3**: Enforcing strong $\varnothing$IC of a WCSP

---

The procedure `enforceS∅IC()` is correct and must terminate. Its complexity can be analyzed by abstracting the worst-case time complexities of `find∅Support()` and `enforce∅IC()` as $f_{strong}$ and $f_{\varnothing IC}$ respectively. Using an augment similar to the proof of Larrosa and Schiex's (2004) Theorems 12 and 21, the complexity can be stated as follows.

**Theorem 2** *The procedure* `enforceS∅IC()` *a time complexity of* $O(r^2 edf_{strong} + ndf_{\varnothing IC})$, *where* $r$ *is the maximum arity of all cost functions,* $d$ *is maximum domain size,* $e = |\mathcal{C}^+|$ *and* $n = |\mathcal{X}|$.

**Proof:** The while loop at line 2 iterates at most $O(nd)$ times. In each iteration, line 6 executes at most $O(r \cdot |N(j)|)$ times, where $N(j)$ is the set of soft constraints restricting $x_j$. Since line 7 executes at most $O(nd)$ times, the overall time complexity is $O(rdf_{strong} \cdot \sum_{j=1}^{n} |N(j)| + ndf_{\varnothing IC}) = O(r^2 edf_{strong} + ndf_{\varnothing IC})$. $O(\sum_{j=1}^{n} |N(j)|) = O(re)$ holds since each cost function counts at most $r$ times in $\sum_{j=1}^{n} |N(j)|$. Thus, it must terminate. $\square$

**Corollary 1** *The procedure* `enforceS∅IC()` *must terminate. The resultant WCSP is strong* $\varnothing IC$, *and equivalent to the original WCSP.*

In general, due to `enforce∅IC()` and `find∅Support()`, enforcing strong $\varnothing$IC is exponential in $r$. As discussed before, `enforce∅IC()` can be reduced to polynomial time for flow-based global cost functions. Similarly, `find∅Support()` can be executed efficiently and incrementally for flow-based global cost functions since line 10 can be computed in polynomial time using minimum cost flow.

Another property we are interested in is *confluence*. A consistency $\Psi$ is *confluent* if enforcing $\Psi$ always transforms a problem $P$ into a unique problem $P'$ which is $\Psi$. AC* is not confluent (Larrosa & Schiex, 2004). With different variable and/or cost function orderings, AC* enforcement can lead to different equivalent WCSPs with different values of $W_\varnothing$. BAC$^\varnothing$ is confluent (Zytnicki et al., 2009). Following the proofs of Propositions 3.3 and 4.3 by Zytnicki et al., it can be shown that strong $\varnothing$IC is also confluent.

**Theorem 3** *(Confluence) Given a WCSP* $P = (\mathcal{X}, \mathcal{D}, \mathcal{C}, \top)$, *there exists a unique WCSP* $P' = (\mathcal{X}, \mathcal{D}', \mathcal{C}', \top)$ *which is strong* $\varnothing IC$ *and equivalent to* $P$.

The above concludes the theoretical analysis of strong $\varnothing$IC. In the following, we compare the strength of strong $\varnothing$IC with the classical consistency notions used in constraint optimization. Following Petit et al. (2000), we define the *reified form of a WCSP* as follows:

**Definition 11** *(Petit et al., 2000) Given a WCSP* $P = (\mathcal{X}, \mathcal{D}, \mathcal{C}, \top)$. *The* reified form, `reified(P)`, *of* $P$ *is a constraint optimization problem (COP)* $(\mathcal{X}^h, \mathcal{D}^h, \mathcal{C}^h, obj)$, *where:*

- $\mathcal{X}^h = \mathcal{X} \cup Z$, *where* $Z = \{z_S \mid W_S \in \mathcal{C} \setminus \{W_\varnothing\}\}$ *are the* cost *variables.*

- $D^h(x_i) = D(x_i)$ *for* $x_i \in \mathcal{X}$, *and* $D^h(z_S) = \{0, \ldots, \top - W_\varnothing - 1\}$ *for each* $z_S \in Z$. *If* $\top - W_\varnothing < 1$, $D^h(z_S) = \varnothing$.

- $\mathcal{C}^h$ *contains the* reified *constraints* $C^h_{S \cup \{z_S\}}$, *which are the hard constraints associated with each* $W_S \in \mathcal{C} \setminus \{W_\varnothing\}$ *defined as* $W_S(\ell) \leq z_S$ *for each tuple* $\ell \in \mathcal{L}(S)$. $\mathcal{C}^h$ *also contains* $C^h_Z$ *defined as* $W_\varnothing \oplus \bigoplus_{z_S \in Z} z_S < \top$.

- *The objective is to minimize* $obj$, *where* $obj = W_\varnothing \oplus \bigoplus_{z_S \in Z} z_S$.

Finding the optimal solution of $\mathtt{reified}(P)$ is equivalent to solving $P$. However, enforcing GAC on $\mathtt{reified}(P)$ cannot remove more values than enforcing strong $\varnothing$IC of $P$. It is because strong $\varnothing$IC of $P$ implies GAC of $\mathtt{reified}(P)$ but not vice versa.

In general, we define the strength comparison as follows.

**Definition 12** *Given a problem $P$ representable by two models $\phi(P)$ and $\psi(P)$. A consistency $\Phi$ on $\phi(P)$ is strictly stronger than another consistency $\Psi$ on $\psi(P)$, written as $\Phi$ on $\phi(P) > \Psi$ on $\psi(P)$, or $\Phi > \Psi$ if $\phi(P) = \psi(P)$, iff $\psi(P)$ is $\Psi$ whenever $\phi(P)$ is $\Phi$, but not vice versa.*

Zytnicki et al. (2009) also define consistency strength comparison in terms of unsatisfiability detection, which is subsumed by our new definition. If $\Phi$ on $\phi(P)$ implies $\Psi$ on $\psi(P)$, and enforcing $\Psi$ on $\psi(P)$ detects unsatisfiability, enforcing $\Phi$ on $\phi(P)$ can detect unsatisfiability as well.

Given a WCSP $P = (\mathcal{X}, \mathcal{D}, \mathcal{C}, \top)$. We show strong $\varnothing$IC on $P$ is stronger than GAC on $\mathtt{reified}(P)$ by the following theorem.

**Theorem 4** *Strong $\varnothing$IC on $P > $ GAC on $\mathtt{reified}(P)$.*

**Proof:** Figure 2 has given an example that a WCSP whose reified COP is GAC may not be strong $\varnothing$IC. We have to show that strong $\varnothing$IC on $P$ implies GAC on $\mathtt{reified}(P)$.

First, $C_Z^h$ is GAC. If $|\mathcal{C}| \leq 1$, the constraint is obviously GAC. If $|\mathcal{C}| > 1$, for each $v_{S_i} \in D(z_{S_i})$, to satisfy the constraint, we just let other cost variables take the value 0, *i.e.* supports for each $v_{S_i} \in D(z_{S_i})$ exist.

Besides, $C_{S \cup \{z_S\}}^h$ is GAC. By the definition of $\varnothing$IC, there exists a tuple $\ell' \in \mathcal{L}(S)$ such that $W_S(\ell') = 0$. The tuple $\ell'$ can form the support of $v_S \in D(z_S)$ with respect to $C_{S \cup \{z_S\}}^h$. Besides, the $\varnothing$-support $\ell_\varnothing$ of $v \in D(x_i)$, together with $v_S = W_S(\ell_\varnothing)$, forms a support for $v \in D(x_i)$. □

For a detailed comparison between strong $\varnothing$IC of WCSPs and GAC of the reified approach, readers can refer to the work of Leung (2009).

When the cost functions are binary, strong $\varnothing$IC cannot be stronger than AC*. In the next section, we show this fact by proving *GAC\**, a generalized version of AC*, to be stronger than strong $\varnothing$IC.

### 4.2 Generalized Arc Consistency

**Definition 13** *(Cooper & Schiex, 2004) Given a WCSP $P = (\mathcal{X}, \mathcal{D}, \mathcal{C}, \top)$. Consider a cost function $W_S \in \mathcal{C}^+$ and a variable $x_i \in S$. A tuple $\ell \in \mathcal{L}(S)$ is a simple support of $v \in D(x_i)$ with respect to $W_S$ with $x_i \in S$ iff $\ell[x_i] = v$ and $W_S(\ell) = 0$. A variable $x_i \in S$ is star generalized arc consistent (GAC\*) with respect to $W_S$ iff $x_i$ is NC\*, and each value $v_i \in D(x_i)$ has a simple support $\ell$ with respect to $W_S$. A WCSP is GAC\* iff all variables are GAC\* with respect to all related non-unary cost functions.*

The definition is designed with practical considerations, and is slightly weaker than Definition 4.2 in the work of Cooper et al. (2010), which also requires $W_S(\ell) = \top$ if $W_\varnothing \oplus \bigoplus_{x_i \in S} W_i(\ell[x_i]) \oplus W_S(\ell) = \top$.

GAC* collapses to AC* for binary cost functions (Larrosa & Schiex, 2004) and AC for ternary cost functions (Sanchez et al., 2008). GAC* is stronger than strong $\varnothing$IC, as a WCSP which is GAC* is also strong $\varnothing$IC, but not vice versa. We state without proof as follows.

**Theorem 5** *GAC\* $>$ strong $\varnothing$IC.*

The procedure `enforceGAC*()` in Algorithm 4 enforces GAC* for a WCSP $(\mathcal{X}, \mathcal{D}, \mathcal{C}, \top)$, based on the `W-AC*3()` Algorithm (Larrosa & Schiex, 2004). The propagation queue $\mathbb{Q}$ stores a set of variables $x_j$. If $x_j \in \mathbb{Q}$, all variables involved in the same cost functions as $x_j$ are potentially not GAC*. Initially, all variables are in $\mathbb{Q}$. A variable $x_j$ is pushed into $\mathbb{Q}$ only after values are removed from $D(x_j)$. At each iteration, an arbitrary variable $x_j$ is removed from the queue by the function `pop()` at line 4. The function `findSupport()` at line 7 enforces GAC* of $x_i$ with respect to $W_S$ by finding the simple supports. The infeasible values are removed by the function `pruneVal()` at line 10. If a value is removed from $D(x_i)$, the simple supports of other related variables may be destroyed. Thus, $x_i$ is pushed back to $\mathbb{Q}$ again by the procedure `pruneVal()`. If `GAC*()` terminates, all values in each variable domain must have a simple support. The WCSP is now GAC*.

**Procedure** `enforceGAC*()`
1    $\mathbb{Q} := \mathcal{X}$;
2    $GAC*()$;

**Procedure** `GAC*()`
3    **while** $\mathbb{Q} \neq \varnothing$ **do**
4       $x_j := pop(\mathbb{Q})$;
5       **foreach** $W_S \in \mathcal{C}^+$ *s.t.* $x_j \in S$ **do**
6          **foreach** $x_i \in S \setminus \{x_j\}$ **do**
7             **if** $findSupport(W_S, x_i)$ **then**
                 // For further consistency enforcement. Assume
                    initially empty if not specified
8                $\mathbb{S} := \mathbb{S} \cup \{x_i\}$;
9                $\mathbb{R} := \mathbb{R} \cup \{x_i\}$;
10    $pruneVal()$;

**Function** `findSupport($W_S, x_i$)`
11    $flag := $ **false**;
12    **foreach** $v \in D(x_i)$ **do**
13       $\alpha := min\{W_S(\ell) \mid \ell[x_i] = v\}$;
14       **if** $W_i(v) = 0 \wedge \alpha > 0$ **then** $flag := $ **true**;
15       $W_i(v) := W_i(v) \oplus \alpha$;
16       **foreach** $\ell \in \mathcal{L}(S)$ *s.t.* $\ell[x_i] = a$ **do** $W_S(\ell) := W_S(\ell) \ominus \alpha$;
17    $unaryProject(x_i)$;
18    **return** $flag$;

**Algorithm 4**: Enforcing GAC* for a WCSP

The procedure `enforceGAC*()` in Algorithm 4 is correct and must terminate. The proof is similar to that of Theorem 2. By replacing $f_{strong}$ by $f_{GAC}$ (the worst-case time complexities of `findSupport()`) and $f_{\varnothing IC}$ by $O(nd)$ (the complexity of `pruneVal()`), the complexity of Algorithm 4 can be stated as follows.

**Theorem 6** *The procedure* `enforceGAC*()` *has a time complexity of* $O(r^2 edf_{GAC} + n^2 d^2)$, *where* $n$, $d$, $e$, *and* $r$ *are as defined in Theorem 2.*

**Corollary 2** *The procedure* `enforceGAC*()` *must terminate. The resultant WCSP is GAC\*, and equivalent to the original WCSP.*

In general, the procedure `enforceGAC*()` is exponential in the maximum arity of the cost function due to `findSupport()`. The function `findSupport()` consists of two operations: (1) finding the minimum cost of the tuple associated with $\{x_i \mapsto v\}$ at line 13, and (2) performing projection at lines 15 and 16. The time complexity of the first operation is polynomial for a flow-based global cost function $W_S$. The method introduced by van Hoeve et al. (2006) can be applied to the first operation as discussed in Section 4.1. However, the second operation modifies $W_S$ to $W_S'$, which requires changing the costs of an exponential number of tuples. Cooper and Schiex (2004) use a similar technique as the one by Zytnicki et al. (2009) (similar to the technique described in Section 4.1) to make the modification constant time. However, the resulting $W_S'$ may not be flow-based, affecting the time complexity of the subsequent procedure calls. To resolve the issue, we introduce *flow-based projection-safety*. If $W_S$ is flow-based projection-safe, the flow property can be maintained throughout enforcement.

**Definition 14** *Given a property $\mathcal{T}$. A global cost function $W_S$ is $\mathcal{T}$ projection-safe iff $W_S$ satisfies the property $\mathcal{T}$, and for all $W_S'$ derived from $W_S$ by a series of projections and extensions, $W_S'$ also satisfies $\mathcal{T}$.*

In other words, a $\mathcal{T}$ projection-safe cost function $W_S$ still satisfies $\mathcal{T}$ after any numbers of projections or extensions. This facilitates the use of $\mathcal{T}$ to derive efficient consistency enforcement algorithms. In the following, we consider a special form of $\mathcal{T}$ projection-safety, when $\mathcal{T}$ is the flow-based property.

In the following, we first define $\mathcal{FB}$, and show that $\mathcal{FB}$ is the sufficient condition of flow-based projection-safety.

**Definition 15** *A global cost function satisfies $\mathcal{FB}$ if:*

1. *$W_S$ is flow-based, with the corresponding network $G = (V, E, w, c, d)$ with a fixed source $s \in V$ and a fixed destination $t \in V$;*

2. *there exists a subjective function mapping each maximum flow $f$ in $G$ to each tuple $\ell_f \in \mathcal{L}(S)$, and;*

3. *there exists an injection mapping from an assignment $\{x_i \mapsto v\}$ to a subset of edges $\bar{E} \subseteq E$ such that for all maximum flow $f$ and the corresponding tuple $\ell_f$, $\sum_{e \in \bar{E}} f_e = 1$ whenever $\ell_f[x_i] = v$, and $\sum_{e \in \bar{E}} f_e = 0$ whenever $\ell_f[x_i] \neq v$*

**Lemma 1** *Given $W_S$ satisfying $\mathcal{FB}$. Suppose $W_S'$ is obtained from* `Project(`$W_S$`,`$W_i$`,`$v$`,`$\alpha$`)` *or* `Extend(`$W_S$`,`$W_i$`,`$v$`,`$\alpha$`)`*. Then $W_S'$ also satisfies $\mathcal{FB}$.*

**Proof:** We only prove the part for projection, since the proof for extension is similar. We first show that $W_S'$ is flow-based (condition 1). Assume $G = (V, E, w, c, d)$ is the corresponding flow network of $W_S$. After the projection, $G$ can be modified to $G' = (V, E, w', c, d)$, where $w'(e) = w(e) - \alpha$ if $e \in \bar{E}$ is an edge corresponding to $\{x_i \mapsto v\}$ and $w'(e) = w(e)$ otherwise. The resulting $G'$ is the corresponding flow network of $W'_S$, since for the maximum flow $f$ in $G$ with minimum cost:

$$
\begin{aligned}
\sum_{e \in E} w'_e f_e &= \sum_{e \in E} w_e f_e - \alpha \sum_{e \in \bar{E}} f_e \\
&= \min\{W_S(\ell) \mid \ell \in \mathcal{L}(S)\} - \alpha \sum_{e \in \bar{E}} f_e \\
&= \min\{W'_S(\ell) \mid \ell \in \mathcal{L}(S)\}.
\end{aligned}
$$

Moreover, since the topology of $G' = (V, E, w', c, d)$ is the same as that of $G = (V, E, w, c, d)$, $W'_S$ also satisfies conditions 2 and 3. ∎

**Theorem 7** *If a global cost function $W_S$ satisfies $\mathcal{FB}$, then $W_S$ is flow-based projection-safe.*

**Proof:** Initially, if no projection and extension is performed, directly from Definition 15, $W_S$ is flow-based. Assume $W'_S$ is the cost function formed from $W_S$ after a series of projections and/or extensions. By Lemma 1, $W'_S$ still satisfies $\mathcal{FB}$ and thus flow-based. Result follows. ∎

As shown by Theorem 7, if a global cost function is flow-based projection-safe, it is always flow-based after projections and/or extensions. Besides, by checking the conditions in Definition 15, we can determine whether a global cost function is flow-based projection-safe.

Note that the computation in the proof is performed under the standard integer set instead of $V(\top)$ for practical considerations. Further investigation is required if the computation can be restricted on $V(\top)$.

By using Theorem 7, we can apply the results by van Hoeve et al. (2006) to compute the value $\min\{W_S(\ell) \mid \ell[x_i] = v \wedge \ell \in \mathcal{L}(S)\}$ in polynomial time throughout GAC* enforcement. Besides, the proof gives an efficient algorithm to perform projection in polynomial time by simply modifying the weights of the corresponding edges.

Again, we use SOFT_ALLDIFFERENT$^{dec}$ as an example. Van Hoeve et al. (2006) have shown that SOFT_ALLDIFFERENT$^{dec}(S)$ satisfies conditions 1 and 2 in Definition 15. Besides, from the network structure shown in Figure 1, by taking $\bar{E} = \{(x_i, v)\}$ for each assignment $\{x_i \mapsto v\}$, condition 3 can be satisfied. Thus, SOFT_ALLDIFFERENT$^{dec}$ is flow-based projection-safe.

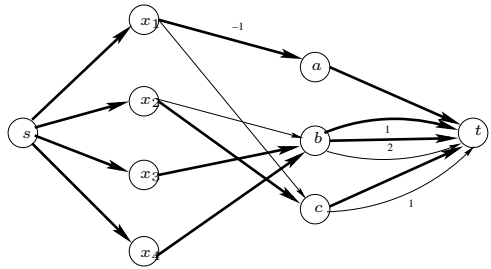

Figure 3: The flow network SOFT_ALLDIFFERENT$^{dec}()$ after projection

Consider the flow network of the SOFT_ALLDIFFERENT$^{dec}$ in Figure 1. Suppose we perform `Project(SOFT_ALLDIFFERENT`$^{dec}$`(S),W`$_1$`,a,1)`. The network is modified to the one in Figure 3, the weight of the edge $(x_1, a)$ in which is decreased from 0 to $-1$. The flow has a cost of 0, which is the cost of the tuple $(a, c, b, b)$ after projection.

If a global cost function is flow-based projection-safe, `findSupport()` has a time complexity depending on the time complexity of computing the minimum cost flow and the shortest path from any two nodes in the network. The result is stated by the following theorem.

**Theorem 8** *Given the time complexities of computing the minimum cost flow and the shortest path are $\mathcal{K}$ and $\mathcal{SP}$ respectively. If $W_S$ is flow-based projection-safe, `findSupport()` has a time complexity of $O(\mathcal{K} + \varepsilon d \cdot \mathcal{SP})$, where $d = max\{|D(x_i)| \mid x_i \in S\}$ and $\varepsilon$ is the maximum size of $\bar{E}$.*

**Proof:** By Theorem 1, after finding a first flow by $O(\mathcal{K})$, the minimum cost at line 13 can be found by augmenting the existing flow, which only requires $O(\mathcal{SP})$. Line 15 can be done in constant time, while line 16 can be done as follows: (a) decrease the weights of all edges corresponding to $x_i \mapsto v$ by $\alpha$, and (b) augment the current flow to the one with new minimum cost by changing the flow values of the edges whose weights have been modified in the first step. The first step requires $O(\varepsilon)$, while the second step requires $O(\varepsilon \cdot \mathcal{SP})$. At most $\varepsilon$ edges are required to change their flow values to maintain minimality of the flow cost. Since `unaryProject()` requires $O(d)$, the overall time is $O(\mathcal{K} + d(\mathcal{SP} + \varepsilon \cdot \mathcal{SP}) + d) = O(\mathcal{K} + \varepsilon d \cdot \mathcal{SP})$. □

The time complexity for finding a shortest path in a graph $\mathcal{SP}$ varies by applying different algorithms. In general, $\mathcal{SP} = O(|V||E|)$, as negative weights are introduced in the graph. However, it can be reduced by applying a potential value on each vertices, as in Johnson's (1977) algorithm. For example, in Figure 3, we can increase the potential value of vertices $a$ and $t$ by 1, and the weight of the edges $(b, t)$ and $(c, t)$ by 1. This increases the cost of all paths from $s$ to $t$ by 1, and makes the weights of all edges non-negative. Dijkstra's (1959) algorithm can thus be applied, reducing the time complexity to $O(|E| + |V|log(|V|))$.

Although GAC* can be enforced in polynomial time for flow-based projection-safe global cost functions, the `findSupport()` function still requires runtime much higher than that for binary or ternary table cost functions in general. To optimize the performance of the solver, we can delay the consistency enforcement of global cost functions until all binary or ternary table cost functions are processed at line 5.

FDAC* for binary cost functions (Larrosa & Schiex, 2003) suggests that a stronger consistency can be deduced by using the extension operator. We will discuss the generalized version of FDAC* for non-binary cost functions in the next section.

### 4.3 Full Directional Generalized Arc Consistency

**Definition 16** *Given a WCSP $P = (\mathcal{X}, \mathcal{D}, \mathcal{C}, \top)$. Consider a cost function $W_S \in \mathcal{C}^+$ and a variable $x_i \in S$. A tuple $\ell$ is the* full support *of the value $v \in D(x_i)$ with respect to $W_S$ and a subset of variables $U \subseteq S \setminus \{x_i\}$ iff $\ell[x_i] = v$ and $W_S(\ell) \oplus \bigoplus_{x_j \in U} W_j(\ell[x_j]) = 0$. A variable $x_i$ is* directional star generalized arc consistent *(DGAC*) with respect to $W_S$ if it is NC* and each value $v \in D(x_i)$ has a full support with respect to $\{x_u \mid x_u \in S \wedge u > i\}$. A WCSP is* full directional star generalized arc consistent *(FDGAC*) if it is GAC* and each variable is DGAC* with respect to all related non-unary cost functions.*

FDGAC* collapses to GAC when WCSPs collapse to CSPs. Moreover, FDGAC* collapses to FDAC* (Larrosa & Schiex, 2003) when the arity of the cost functions is two. However, FDGAC* is incomparable to FDAC for ternary cost functions (Sanchez et al., 2008). FDAC requires full supports with not only zero unary but also zero binary costs for the next variable in $S$ only, while we only require all variables with full supports of zero unary costs.

By definition, FDGAC* is stronger than GAC* and also strong $\varnothing$IC.

**Theorem 9** *FDGAC\* > GAC\* > strong $\varnothing$IC.*

The procedure `enforceFDGAC*()` enforces FDGAC* for a WCSP, based on the `FDAC*()` Algorithm (Larrosa & Schiex, 2003). The propagation queues $\mathbb{Q}$ and $\mathbb{R}$ store a set of variables. If $x_j \in \mathbb{Q}$, all variables involved in the same cost functions as $x_j$ are potentially not GAC*; if $x_j \in \mathbb{R}$, the variables $x_i$ involved in the same cost functions as $x_j$ are potentially not DGAC*. When values are removed from the domain of variable $x_j$, $x_j$ is pushed onto $\mathbb{Q}$ and $\mathbb{R}$; when unary costs of the values in $D(x_j)$ are increased, $x_j$ is pushed to $\mathbb{R}$. At each iteration, GAC* is maintained by the procedure `GAC*()`. DGAC* is then enforced by `DGAC*()`. Enforcing DGAC* follows the ordering from the largest index to the smallest index such that the full supports of values in the domains of variables with smaller indices are not destroyed by DGAC*-enforcement on those with larger indices. The variable with the largest index in $\mathbb{R}$ is removed from $\mathbb{R}$ by the function `popMax()`. By implementing $\mathbb{R}$ as a heap, `popMax()` requires only constant time. DGAC* enforcement is performed at line 10 by `findFullSupport()`. In the last step, NC* is re-enforced by `pruneVal()`. The iteration continues until all propagation queues are empty, which implies all values in each variable domain has a simple and full support, and all variables are NC*. The resultant WCSP is FDGAC*.

---

**Procedure** `enforceFDGAC*()`

1    $\mathbb{R} := \mathbb{Q} := \mathcal{X}$;
2    **while** $\mathbb{R} \neq \varnothing \vee \mathbb{Q} \neq \varnothing$ **do**
3      $GAC\star$ ();
4      $DGAC\star$ ();
5      $pruneVal$ ();

**Procedure** `DGAC*()`

6    **while** $\mathbb{R} \neq \varnothing$ **do**
7      $x_u := popMax$ ($\mathbb{R}$);
8      **foreach** $W_S \in \mathcal{C}^+$ *s.t.* $x_u \in S$ **do**
9        **for** $i = n$ **DownTo** 1 *s.t.* $x_i \in S \setminus \{x_u\}$ **do**
10          **if** $findFullSupport$ ($W_S, x_i, S \cap \{x_j \mid j > i\}$) **then** $\mathbb{R} := \mathbb{R} \cup \{x_i\}$;
12          $\mathbb{S} := \mathbb{S} \cup \{x_i\}$ ;      // For further consistency enforcement.

**Function** `findFullSupport`$(W_S, x_i, U)$

13    **foreach** $x_j \in U$ **do**
14      **foreach** $v_j \in D(x_j)$ **do**
15        **foreach** $\ell \in \mathcal{L}(S)$ *s.t.* $\ell[x_j] = v_j$ **do** $W_S(\ell) := W_S(\ell) \oplus W_j(v_j)$;
16        $W_j(v_j) := 0$;
17    $flag := findSupport$ ($W_S, x_i$);
18    **foreach** $x_j \in U$ **do** $findSupport$ ($W_S, x_j$);
19    $unaryProject$ ($x_i$);
20    **return** $flag$;

**Algorithm 5**: Enforcing FDGAC* on a WCSP

---

The procedure `enforceFDGAC*()` in Algorithm 5 is correct and must terminate, the proof of which is similar to those of Theorems 3 and 4 by Larrosa and Schiex (2003). The worst-case time complexity of `enforceFDGAC*()` can be stated in terms of that of `findFullSupport()` ($f_{DGAC}$) and `findSupport()` ($f_{GAC}$) as follows.

**Theorem 10** *The procedure* `enforceFDGAC*()` *has a time complexity of* $O(r^2 ed(nf_{DGAC} + f_{GAC}) + n^2 d^2)$, *where* n, d, e, *and* r *are as defined in Theorem 2.*

**Proof:** First we analyze the time complexity of enforcing DGAC*. Consider the procedure `DGAC*()` at line 6. The while-loop iterates at most $O(n)$ times. Since no value is removed in the while-loop, once $x_i$ is processed at line 10, where $i > j$, it is not pushed back to $\mathbb{R}$ at line 11. Thus, line 10 executes at most $O(r \sum_{j=0}^n |N(j)|) = O(r^2 e)$ times, where $N(j)$ is the set of cost functions restricting $x_j$. Therefore, the time complexity of `DGAC*()` is $O(r^2 ef_{DGAC})$. Since `DGAC*()` executes at most $O(nd)$ times throughout the global enforcement iteration. Thus the time spent on enforcing DGAC* is $O(nr^2 edf_{DGAC})$

Although `GAC*()` is called $O(nd)$ times, it does nothing if no values are removed from variable domains. Thus we count the number of times calling `findSupport()`. Since the variables are pushed into $\mathbb{Q}$ only when a value is removed, `findSupport()` only executes at most $O(nd)$ times throughout the global enforcement iteration. Similar arguments apply to `pruneVal()` at line 10 inside `GAC*()` defined in Algorithm 4. With the proof similar to Theorem 6, the time spent on enforcing GAC* is $O(r^2 edf_{GAC} + n^2 d^2)$.

The `pruneVal` at line 5 executes $O(nd)$ times, and each time it requires a time complexity of $O(nd)$. Therefore, the overall time complexity is $O(r^2 ed(nf_{DGAC} + f_{GAC}) + n^2 d^2)$. □

**Corollary 3** *The procedure* `enforceFDGAC*()` *must terminate. The resultant WCSP is FDGAC\** *and equivalent to the original WCSP.*

Again, the complexity is exponential in the maximum arity due to the function `findSupport()` and `findFullSupport()`. In the following, we focus the discussion on `findFullSupport()`. The first part (lines 15 and 16) performs extensions to push all the unary costs back to $W_S$. By the time we execute line 17, all unary costs $W_j$, where $x_j \in U$, are 0, and enforcing GAC* for $x_i$ achieves the second requirement of DGAC* (each $v \in D(x_i)$ has a full support). Line 18 re-instates GAC* for all variables $x_j \in U$. Note that success in line 17 guarantees that $W_j(v_j) = 0$ for some value $v_j$ appearing in a tuple $\ell$ which makes $W_S(\ell) = 0$.

Again, flow-based projection-safety helps reduce the time complexity of `findFullSupport()` throughout the enforcement. The proof of Theorem 7 gives a polynomial time algorithm to perform extension and maintain efficient computation of $\min\{W_S(\ell) \mid \ell \in \mathcal{L}(S)\}$. Flow-based projection-safety can be guaranteed by Theorem 7, which requires checking conditions 1, 2, and 3 in the definition of flow-based projection-safety. The complexity result follows from Theorems 2 and 8.

**Theorem 11** *If* $W_S$ *is a flow-based projection-safe global cost function,* `findFullSupport()` *has a time complexity of* $O(\mathcal{K} + \varepsilon rd \cdot \mathcal{SP})$, *where* r, $\varepsilon$, d, $\mathcal{K}$ *and* $\mathcal{SP}$ *are as defined in Theorems 2 and 8.*

**Proof:** Similarly to Theorem 8, lines 13 to 16 can be performed as follows: (a) for each $x_j \in U$ and each value $v_j \in D(x_j)$, increase the weights of all edges corresponding to $\{x_j \mapsto v_j\}$ by $W_j(v_j)$, and then reduce $W_j(v_j)$ to 0, and (b) find a flow with the new minimum cost in the new flow network. The first step can be done in $O(\varepsilon rd)$, as the size of $U$ is bounded by the arity of the cost function $r$. The second step can be done in $O(\mathcal{K})$, which also acts as preprocessing for `findSupport()` at lines 17 and 18. By Theorem 8, lines 17 and 18 can be done in $O(r\varepsilon d \cdot \mathcal{SP})$. Thus, the overall complexity is $O(r \cdot \varepsilon d + \mathcal{K} + r\varepsilon d \cdot \mathcal{SP}) = O(\mathcal{K} + \varepsilon rd \cdot \mathcal{SP})$. □

Similarly to GAC*, the DGAC* enforcement for global cost functions can be delayed until all binary and ternary table cost functions are processed.

## 4.4 Generalizing Existential Directional Arc Consistency

EDAC* (de Givry et al., 2005) can be generalized to EDGAC* using the full support definition as in FDGAC*. However, we find that naively generalizing EDAC* is not always enforceable, due to the limitation of EDAC*. In the following, we explain and provide a solution to this limitation.

### 4.4.1 AN INHERENT LIMITATION OF EDAC*

**Definition 17** *(de Givry et al., 2005) Consider a binary WCSP $P = (\mathcal{X}, \mathcal{D}, \mathcal{C}, \top)$. A variable $x_i \in \mathcal{X}$ is* existential arc consistent *(EAC*) if it is NC* and there exists a value $v \in D(x_i)$ with zero unary cost such that it has full supports with respect to all binary cost functions $W_{i,j}$ on $\{x_i, x_j\}$ and $\{x_j\}$. $P$ is* existential directional arc consistent *(EDAC*) if it is FDAC* and all variables are EAC*.*

Enforcing EAC* on a variable $x_i$ requires two main operations: (1) compute

$$\alpha = \min_{a \in D(x_i)}\{W_i(a) \oplus \bigoplus_{W_{i,j} \in \mathcal{C}} \min_{b \in D(x_j)}\{W_{i,j}(a,b) \oplus W_j(b)\}\},$$

which determines whether enforcing full supports breaks the NC* requirement, and (2) if $\alpha > 0$, enforce full supports with respect to all cost functions $W_{i,j} \in \mathcal{C}$ by invoking `findFullSupport($x_i$, $W_{i,j}$, $\{x_j\}$)`, implying that NC* is no longer satisfied and hence $W_\varnothing$ can be increased by enforcing NC*. EDAC* enforcement will oscillate if constraints share more than one variable. The situation is similar to Example 3 by de Givry et al. (2005). We demonstrate by the example in Figure 4(a), which shows a WCSP with two cost functions $W^1_{1,2}$ and $W^2_{1,2}$. It is FDAC* but not EDAC*. If $x_2$ takes the value $a$, $W^1_{1,2}(v,a) \oplus W_1(v) \geq 1$ for all values $v \in D(x_1)$; if $x_2$ takes the value $b$, $W^2_{1,2}(v,b) \oplus C_1(v) \geq 1$ for all values $v \in D(x_1)$. Thus, by enforcing full supports of each value in $D(x_2)$ with respect to all cost functions and $\{x_1\}$, NC* is broken and $W_\varnothing$ can be increased. To increase $W_\varnothing$, we enforce full supports: the cost of 1 in $W_1(a)$ is extended to $W^1_{1,2}$, resulting in Figure 4(b). No costs in $W_1$ can be extended to $W^2_{1,2}$. Performing projection from $W^1_{1,2}$ to $W_2$ results in Figure 4(c). The WCSP is now EAC* but not FDAC*. Enforcing FDAC* converts the problem state back to Figure 4(a).

The problem is caused by the first step, which does not tell how the unary costs are separated for extension to increase $W_\varnothing$. Although an increment is predicted, the unary cost in $W_1(a)$ has a choice of moving itself to $W^1_{1,2}$ or $W^2_{1,2}$. During computation, no information is obtained on how the unary costs are moved. As shown, a wrong movement breaks DAC* without incrementing $W_\varnothing$, resulting in oscillation.

This problem does not occur in existing solvers which handle only up to ternary cost functions. The solvers allow only one binary cost functions for every pair of variables. If there are indeed two cost functions for the same two variables, the cost functions can be merged into one, where the

**(a) Original WCSP** — $\top = 4, W_\varnothing = 0$

| $x_1$ | $W_1$ |
|---|---|
| $a$ | 1 |
| $b$ | 0 |

| $x_1$ | $x_2$ | $W^1_{1,2}$ |
|---|---|---|
| $a$ | $a$ | 0 |
| $a$ | $b$ | 2 |
| $b$ | $a$ | 1 |
| $b$ | $b$ | 0 |

| $x_2$ | $W_2$ |
|---|---|
| $a$ | 0 |
| $b$ | 0 |

| $x_1$ | $x_2$ | $W^2_{1,2}$ |
|---|---|---|
| $a$ | $a$ | 1 |
| $a$ | $b$ | 0 |
| $b$ | $a$ | 0 |
| $b$ | $b$ | 2 |

**(b) After Extension** — $\top = 4, W_\varnothing = 0$

| $x_1$ | $W_1$ |
|---|---|
| $a$ | $\underline{0}$ |
| $b$ | 0 |

| $x_1$ | $x_2$ | $W^1_{12}$ |
|---|---|---|
| $a$ | $a$ | $\underline{1}$ |
| $a$ | $b$ | $\underline{3}$ |
| $b$ | $a$ | 1 |
| $b$ | $b$ | 0 |

| $x_2$ | $W_2$ |
|---|---|
| $a$ | 0 |
| $b$ | 0 |

| $x_1$ | $x_2$ | $W^2_{12}$ |
|---|---|---|
| $a$ | $a$ | 1 |
| $a$ | $b$ | 0 |
| $b$ | $a$ | 0 |
| $b$ | $b$ | 2 |

**(c) After Projection** — $\top = 4, W_\varnothing = 0$

| $x_1$ | $W_1$ |
|---|---|
| $a$ | 0 |
| $b$ | 0 |

| $x_1$ | $x_2$ | $W^1_{12}$ |
|---|---|---|
| $a$ | $a$ | $\underline{0}$ |
| $a$ | $b$ | 3 |
| $b$ | $a$ | $\underline{0}$ |
| $b$ | $b$ | 0 |

| $x_2$ | $W_2$ |
|---|---|
| $a$ | 1 |
| $b$ | 0 |

| $x_1$ | $x_2$ | $W^2_{12}$ |
|---|---|---|
| $a$ | $a$ | 1 |
| $a$ | $b$ | 0 |
| $b$ | $a$ | 0 |
| $b$ | $b$ | 2 |

Figure 4: Oscillation in EDAC* enforcement

cost of a tuple in the merged function is the sum of the costs of the same tuple in the two original functions. However, if we allow high arity global cost functions, sharing of more than one variable would be common and necessary in many scenarios. A straightforward generalization of EDAC* for non-binary cost functions would inherit the same oscillation problem. In the case of ternary cost functions, Sanchez et al. (2008) cleverly avoid the oscillation problem by re-defining full supports to include not just unary but also binary cost functions. During EDAC enforcement, unary costs are distributed through extension to binary cost functions. However, the method is only designed for ternary cost functions. In the following, we define a weak version of EDAC*, which is based on the notion of *cost-providing partitions*.

### 4.4.2 Cost-Providing Partitions and Weak EDGAC*

**Definition 18** *A* cost-providing partition $\mathcal{B}_{x_i}$ *for variable* $x_i \in \mathcal{X}$ *is a set of sets* $\{B_{x_i,W_S} \mid x_i \in S\}$ *such that:*

- $|\mathcal{B}_{x_i}|$ *is the number of constraints which scope includes* $x_i$;
- $B_{x_i,W_S} \subseteq S$;
- $B_{x_i,W_{S_j}} \cap B_{x_i,W_{S_k}} = \varnothing$ *for any two different constraints* $W_{S_k}, W_{S_j} \in \mathcal{C}^+$, *and;*
- $\bigcup_{B_{x_i,W_S} \in \mathcal{B}_{x_i}} B_{x_i,W_S} = (\bigcup_{W_S \in \mathcal{C}^+ \wedge x_i \in S} S) \setminus \{x_i\}$.

Essentially, $\mathcal{B}_{x_i}$ forms a partition of the set containing all variables constrained by $x_i$. If $x_j \in B_{x_i,W_S}$, the unary costs in $W_j$ can only be extended to $W_S$ when enforcing EAC* for $x_i$. This avoids the problem of determining how the unary costs of $x_j$ are distributed when there exists more than one constraint on $\{x_i, x_j\}$.

Based on the cost-providing partitions, we define *weak EDAC*.

**Definition 19** *Consider a binary WCSP* $P = (\mathcal{X}, \mathcal{D}, \mathcal{C}, \top)$ *and cost-providing partitions* $\{\mathcal{B}_{x_i} \mid x_i \in \mathcal{X}\}$. *A* weak fully supported value $v \in D(x_i)$ *of a variable* $x_i \in \mathcal{X}$ *is a value with zero unary cost and for each variable* $x_j$ *and a binary cost function* $W^m_{i,j}$, *there exists a value* $b \in D(x_j)$ *such that* $W^m_{i,j}(v,b) = 0$ *if* $B_{x_i,W^m_{i,j}} = \{\}$, *and* $W^m_{i,j}(v,b) \oplus W_j(b) = 0$ *if* $B_{x_i,W^m_{i,j}} = \{x_j\}$. *A variable* $x_i$ *is* weak existential arc consistent *(weak EAC*) if it is NC* and there exists at least one weak fully supported value in its domain.* $P$ *is* weak existential directional arc consistent *(weak EDAC*) if it is FDAC* and each variable is weak EAC*.*

Weak EDAC* collapses to AC when WCSPs collapse to CSPs for any cost-providing partition. Moreover, weak EDAC* is reduced to EDAC* (de Givry et al., 2005) when the binary cost functions share at most one variable.

We further generalize weak EDAC* to *weak EDGAC** for $n$-ary cost functions.

**Definition 20** *Given a WCSP $P = (\mathcal{X}, \mathcal{D}, \mathcal{C}, \top)$ and cost-providing partitions $\{\mathcal{B}_{x_i} \mid x_i \in \mathcal{X}\}$. A weak fully supported value $v \in D(x_i)$ of a variable $x_i$ is a value with zero unary cost and full supports with respect to all cost functions $W_S \in \mathcal{C}^+$ with $x_i \in S$ and $B_{x_i, W_S}$. A variable $x_i$ is* weak existential generalized arc consistent *(weak EGAC*) if it is NC* and there exists at least one weak fully supported value in its domain. $P$ is* weak existential directional generalized arc consistent *(weak EDGAC*) if it is FDGAC* and each variable is weak EGAC*.*

Weak EDAC* and weak EDGAC* can be achieved using for any cost-providing partitions. Weak EDGAC* is reduced to GAC when WCSPs collapse to CSPs.

Compared with other consistency notions, weak EDGAC* is strictly stronger than FDGAC* and other consistency notions we have described. It can be deduced directly from the definition.

**Theorem 12** *For any cost-providing partitions, weak EDGAC* > FDGAC* > GAC* > strong $\varnothing$IC*

VAC is stronger than weak EDGAC*, as stated in the theorem below.

**Theorem 13** *VAC are strictly stronger than weak EDGAC* with any cost-providing partition.*

**Proof:** A WCSP which is VAC must be weak EDGAC* for any cost-providing partition. Otherwise, there must exist a sequence of projections and extensions to increase $W_\varnothing$, which violates Theorem 7.3 by Cooper et al. (2010). On another hand, Cooper et al. (2010) give an example which is EDAC* but not VAC. Results follow. □

However, weak EDGAC* is incomparable to complete $k$-consistency (Cooper, 2005), where $k > 2$, for any cost-providing partition. It is because EDAC* is already incomparable to complete $k$-consistency (Sanchez et al., 2008).

To compute the cost-providing partition $\mathcal{B}_{x_i}$ of a variable $x_i$, we could apply Algorithm 6, which is a greedy approach to partition the set $Y$ containing all variables related to $x_i$ defined in line 1, hoping to gathering more costs by gathering more variables at one cost function, increasing the chance of removing more infeasible values and raising $W_\varnothing$.

---

**Procedure** `findCostProvidingPartition`($x_i$)

1   $Y = (\bigcup_{W_S \in \mathcal{C}^+ \wedge x_i \in S} S) \setminus \{x_i\}$;
2   Sort $\mathcal{C}^+$ in decreasing order of $|S|$;
3   **foreach** $W_S \in \mathcal{C}^+$ *s.t.* $x_i \in S$ **do**
4      $B_{x_i, W_S} = Y \cap S$;
5      $Y = Y \setminus S$;

**Algorithm 6**: Finding $\mathcal{B}_{x_i}$

---

The procedure `enforceWeakEDGAC*()` in Algorithm 7 enforces weak EDGAC* of a WCSP. The cost-providing partitions are first computed in line 1. The procedure makes use of four propagation queues $\mathbb{P}$, $\mathbb{Q}$, $\mathbb{R}$ and $\mathbb{S}$. If $x_i \in \mathbb{P}$, the variable $x_i$ is potentially not weak EGAC* due to

**Procedure** `enforceWeakEDGAC*()`

1  **foreach** $x_i \in \mathcal{X}$ **do** `findCostProvidingPartition`$(x_i)$;
2  $\mathbb{R} := \mathbb{Q} := \mathbb{S} := \mathcal{X}$;
3  **while** $\mathbb{S} \neq \varnothing \vee \mathbb{R} \neq \varnothing \vee \mathbb{Q} \neq \varnothing$ **do**
4  $\quad$ $\mathbb{P} := \mathbb{S} \cup \bigcup_{x_i \in \mathbb{S}, W_S \in \mathcal{C}^+}(S \setminus \{x_i\})$;
5  $\quad$ `weakEGAC*`();
6  $\quad$ $\mathbb{S} := \varnothing$;
7  $\quad$ `DGAC*`();
8  $\quad$ `GAC*`();
9  $\quad$ `pruneVal`();

**Procedure** `weakEGAC*()`

10  **while** $\mathbb{P} \neq \varnothing$ **do**
11  $\quad$ $x_i := pop(\mathbb{P})$;
12  $\quad$ **if** `findExistentialSupport`$(x_i)$ **then**
13  $\qquad$ $\mathbb{R} := \mathbb{R} \cup \{x_i\}$;
14  $\qquad$ $\mathbb{P} := \mathbb{P} \cup \{x_j \mid x_i, x_j \in W_S, W_S \in \mathcal{C}^+\}$;

**Function** `findExistentialSupport`$(x_i)$

15  $flag :=$ **false**;
16  $\alpha := \min_{a \in D(x_i)}\{W_i(a) \oplus \bigoplus_{x_i \in S, W_S \in \mathcal{C}^+} \min_{\ell[x_i]=a}\{W_S(\ell) \oplus \bigoplus_{x_j \in B_{x_i, W_S}} W_j(\ell[x_j])\}\}$;
17  **if** $\alpha > 0$ **then**
18  $\quad$ $flag :=$ **true**;
19  $\quad$ **foreach** $W_S \in \mathcal{C}^+$ *s.t.* $x_i \in S$ **do** `findFullSupport`$(W_S, x_i, B_{x_i, W_S})$;
20  **return** $flag$;

**Algorithm 7**: Enforcing weak EDGAC*

a change in unary costs or a removal of values in some variables. If $x_j \in \mathbb{R}$, the variables $x_i$ involved in the same cost functions as $x_j$ are potentially not DGAC*. If $x_j \in \mathbb{Q}$, all variables in the same cost functions as $x_j$ are potentially not GAC*. The propagation queue $\mathbb{S}$ helps build $\mathbb{P}$ efficiently. The procedure `weakEGAC*()` enforces weak EGAC* on each variable by the procedure `findExistentialSupport()` in line 12. If `findExistentialSupport()` returns true, a projection has been performed for some cost functions. The weak fully supported values of other variables may be destroyed. Thus, the variables constrained by $x_i$ are pushed back onto $\mathbb{P}$ for revision in line 14. DGAC* and GAC* are enforced by the procedures `DGAC*()` and `GAC*()`. A change in unary cost requires re-examining DGAC* and weak EGAC*, which is done by pushing the variables into the corresponding queues in lines 13 and 14, and lines 11 and 12 in Algorithm 5. In the last step, NC* is enforced by `pruneVal()`. Again, if a value in $D(x_i)$ is removed, GAC*, DGAC* or weak EGAC* may be destroyed, and $x_i$ is pushed into the corresponding queues for re-examination by `pruneVal()` in Algorithm 1. If all propagation queues are empty, all variables are GAC*, DGAC*, and weak EGAC*, i.e. the WCSP is weak EDGAC*.

The algorithm is correct and must terminate. We analyze the time complexity by abstracting the worst-case time complexities of `findSupport()`, `findFullSupport()` and

`findExistentialSupport()` as $f_{GAC}$, $f_{DGAC}$, and $f_{EGAC}$ respectively. The overall time complexity is stated as follows.

**Theorem 14** *The procedure `enforceWeakEDGAC*()` requires $O((nd+\top)(f_{EGAC}+r^2ef_{DGAC}+nd)+r^2edf_{GAC})$, where $n$, $d$, $e$, and $r$ are defined in Theorem 2.*

**Proof:** As line 1 requires only $O(nr)$, we only analyze the overall time complexity spent by each sub-procedure and compute the overall time complexity.

A variable is pushed into $\mathbb{S}$ if a value is removed or weak EGAC* is violated. The former happens $O(nd)$ times, while the latter occurs $O(\top)$ times (each weak EGAC* is violated, $W_\varnothing$ will be increased). Since $\mathbb{P}$ is built on $\mathbb{S}$, `findExistentialSupport()` is executed at most $O(nd+\top)$ times throughout the global enforcement. Thus, the time complexity spent on enforcing weak EGAC* is $O((nd+\top)f_{EGAC})$.

A variable is pushed into $\mathbb{R}$ if either a value is removed, or unary costs are moved by GAC* or weak EGAC* enforcement. Thus, `DGAC*()` is called $O(nd+\top)$ times. Each time `DGAC*()` is called, by Theorem 10, it requires $O(r^2ef_{DGAC})$ for DGAC* enforcement. Thus, the time complexity of enforcing DGAC* is $O((nd+\top)r^2ef_{DGAC})$.

A variable is pushed into $\mathbb{Q}$ only if a value is removed. Thus, `findSupport()` inside the procedure `GAC*()` is called at most $O(nd)$ times throughout the global enforcement. Using the proof similar to Theorem 6, the overall time spent on enforcing GAC* is $O(r^2edf_{GAC}+n^2d^2)$.

The main while-loop in line 3 terminates when all propagation queues are empty. Thus, the main while-loop iterates $O(nd+\top)$ times. The time complexity for re-enforcing NC* by `pruneVal()` at line 9 is $O((nd+\top)nd)$.

By summing up all time complexity results, the overall time complexity is $O((nd+\top)(f_{EGAC}+r^2ef_{DGAC}+nd)+r^2edf_{GAC})$. $\qquad\square$

**Corollary 4** *The procedure `enforceWeakEDGAC*()` must terminate. The resultant WCSP is weak EDGAC*, and equivalent to the original WCSP.*

The procedure `enforceWeakEDGAC*()` is again exponential due to `findSupport()`, `findFullSupport()` and `findExistentialSupport()`. In the following, we focus on the last procedure. It first checks whether a weak fully supported value exists by computing $\alpha$, which determines whether NC* still holds if we perform `findFullSupport()` from line 19. If $\alpha$ equals 0, a weak fully supported value exists and nothing should be done; otherwise, this value can be made weak fully supported by the for-loop at line 19. The time complexity depends on two operations: (1) computing the value of $\alpha$ in line 16, and; (2) finding full supports by the line 19. These two operations are exponential in $|S|$ in general. However, if all global cost functions are flow-based projection-safe, the time complexity of the above operations can be reduced to polynomial time.

In the next section, we put theory into practice. We demonstrate our framework with different benchmarks and compare the results with the current approach.

## 5. Towards a Library of Efficient Global Cost Functions

In the previous section, we only show SOFT_ALLDIFFERENT$^{dec}$ is flow-based projection-safe. In the following, we further show that a range of common global cost functions are also flow-based projection-safe. We give experimental results on various benchmarks with different consistency notions and different global cost functions.

### 5.1 A List of Flow-Based Projection-Safe Global Cost Functions

In this section, we show that a number of common global cost functions are flow-based projection-safe. They include the soft variants of ALL_DIFFERENT, GCC, SAME, and REGULAR constraints.

#### 5.1.1 The Soft Variants of AllDifferent

The ALLDIFFERENT() constraint restricts variables to take distinct values (Laurière, 1978). There are two possible soft variants, namely SOFT_ALLDIFFERENT$^{dec}$() and ALLDIFFERENT$^{var}$(). The former returns the number of pairs of variables that share the same value, while the latter returns the least number of variables that must be changed so that all variables take distinct values. The cost function SOFT_ALLDIFFERENT$^{dec}$() is shown to be flow-based projection-safe in Section 4.2. In fact, this also implies that another cost function SOFT_ALLDIFFERENT$^{var}$() is flow-based projection-safe. The SOFT_ALLDIFFERENT$^{var}$() function also corresponds to a flow network with structure similar to that of SOFT_ALLDIFFERENT$^{dec}$() but different in weights on the edges connecting to $t$ (van Hoeve et al., 2006). We state the results as follows.

**Theorem 15** *The cost functions* SOFT_ALLDIFFERENT$^{var}$*(S) and* SOFT_ALLDIFFERENT$^{dec}$*(S) are flow-based projection-safe.*

#### 5.1.2 The Soft Variants of GCC

Given a set of values $\Sigma = \bigcup_{x_i \in S} D(x_i)$ and functions $lb$ and $ub$ that maps from $\Sigma$ to non-negative integers. Each value $v \in \Sigma$ is associated with a upper bound $ub_v$ and a lower bound $lb_v$. The GCC$(S, ub, lb)$ constraint is satisfied by a tuple $\ell \in \mathcal{L}(S)$ if the number of occurrences of a value $v \in \Sigma$ in $\ell$ (denoted by $\#(\ell, v)$) is at most $ub_v$ times and at least $lb_v$ times (Régin, 1996). There are two soft variants of GCC constraints, namely SOFT_GCC$^{var}$() and SOFT_GCC$^{val}$() (van Hoeve et al., 2006).

**Definition 21** *(van Hoeve et al., 2006) Define two functions $s(\ell, v)$ and $e(\ell, v)$: $s(\ell, v)$ returns $lb_v - \#(\ell, v)$ if $\#(\ell, v) \leq lb_v$, and 0 otherwise; $e(\ell, v)$ returns $\#(\ell, v) - ub_v$ if $\#(\ell, v) \geq ub_v$, and 0 otherwise.*

*The global cost functions* SOFT_GCC$^{var}$*(S) returns* $\max\{\sum_{v \in \Sigma} s(\ell, v), \sum_{v \in \Sigma} e(\ell, v)\}$*, provided that $\sum_{v \in \Sigma} lb_v \leq |S| \leq \sum_{v \in \Sigma} ub_v$; while* SOFT_GCC$^{val}$*(S) returns* $\sum_{v \in \Sigma}(s(\ell, v) + e(\ell, v))$*.*

Van Hoeve et al. (2006) show that both SOFT_GCC$^{var}$ and SOFT_GCC$^{dec}$ are flow-based, and the flow networks have structures similar to the SOFT_ALLDIFFERENT cost functions. With a proof similar to Theorem 15, we can show the following theorem.

**Theorem 16** *The cost functions* SOFT_GCC$^{var}$*(S) and* SOFT_GCC$^{val}$*(S) are flow-based projection-safe.*

#### 5.1.3 The Soft Variants of Same

Given two sets of variables $S_1$ and $S_2$ with $|S_1| = |S_2|$ and $S_1 \cap S_2 = \varnothing$. The SAME$(S_1, S_2)$ constraint is satisfied by the tuple $\ell \in \mathcal{L}(S_1 \cup S_2)$ if $\ell[S_1]$ is a permutation of $\ell[S_2]$ (Beldiceanu, Katriel, & Thiel, 2004). The hard SAME() constraint can be softened to the global cost function SOFT_SAME$^{var}$() (van Hoeve et al., 2006):

**Definition 22** *(van Hoeve et al., 2006) Given that the union operation $\cup$ is the multi-set union, and $\varphi_1 \Delta \varphi_2$ returns the symmetric difference between two multi-sets $\varphi_1$ and $\varphi_2$, i.e.$\varphi_1 \Delta \varphi_2 = (\varphi_1 \setminus \varphi_2) \cup (\varphi_2 \setminus \varphi_1)$.*

*The global cost function* SOFT_SAME$^{var}(S_1, S_2)$ *returns* $|(\bigcup_{x_i \in S_1} \{\ell[x_i]\}) \Delta (\bigcup_{y_i \in S_2} \{\ell[y_i]\})|/2.$

**Theorem 17** *The cost function* SOFT_SAME$^{var}(S_1, S_2)$ *is flow-based projection-safe.*

**Proof:** Van Hoeve et al. (2006) have shown that SOFT_SAME$^{var}$ satisfies conditions 1 and 2 in Definition 15. For instance, consider $S_1 = \{x_1, x_2, x_3\}$ and $S_2 = \{x_4, x_5, x_6\}$ with $D(x_1) = \{a\}$, $D(x_2) = \{a, b\}$, $D(x_3) = \{b\}$, $D(x_4) = \{a, b\}$, and $D(x_5) = D(x_6) = \{a\}$. The flow network corresponding to SOFT_SAME$^{var}(S_1, S_2)$ is shown in Fig. 5. Solid edges have zero weight and unit capacity. Dotted edges have unit weight and a capacity of 3. The thick edges show the $(s, t)$-flow corresponding to the tuple $\ell = (a, b, b, b, a, a)$.

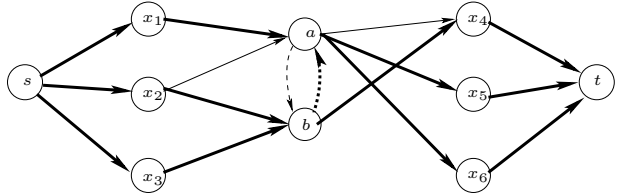

Figure 5: The flow network corresponding to the SOFT_SAME$^{var}(S_1, S_2)$ constraint

Moreover, from the network structure, by taking $\bar{E} = \{(x_i, v)\}$ for $x_i \in S_1$ and $v \in D(x_i)$, and $\bar{E} = \{(v, y_i)\}$ for $y_i \in S_2$ and $v \in D(y_i)$, the cost function satisfies condition 3. Thus, it is flow-based projection-safe. $\square$

### 5.1.4 THE SOFT VARIANTS OF REGULAR

The REGULAR constraint are defined based on regular languages. A regular language $L(M)$ can be represented by a finite state automaton $M = (Q, \Sigma, \delta, q_0, F)$. $Q$ is the set of states. $\Sigma$ is a set of characters. The symbol $q_0 \in Q$ denotes the initial state and $F \subseteq Q$ is the set of final states. The transition function $\delta$ is defined as $\delta : Q \times \Sigma \mapsto Q$. An automaton can be represented graphically as shown in Figure 6, where the final states are denoted by double circles.

Given $D(x_i) \subseteq \Sigma$ for each $x_i \in S$. The REGULAR$(S, M)$ constraint accepts the tuple $\ell \in \mathcal{L}(S)$ if the corresponding string belongs to a regular language $L(M)$ represented by a finite state automaton $M = (Q, \Sigma, \delta, q_0, F)$ (Pesant, 2004).

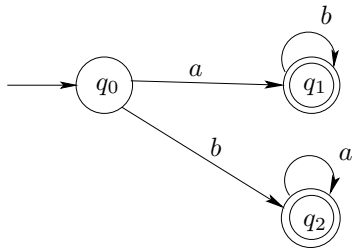

Figure 6: The graphical representation of a automaton.

Two soft variants are defined for the REGULAR constraint, namely SOFT_REGULAR$^{var}$() and SOFT_REGULAR$^{edit}$() (van Hoeve et al., 2006):

**Definition 23** *(van Hoeve et al., 2006) Define $\tau_\ell$ to be the string formed from the tuple $\ell \in \mathcal{L}(S)$. The cost functions SOFT_REGULAR$^{var}(S)$ returns $\min\{H(\tau_\ell, \tau) \mid \tau \in L(M)\}$, where $H(\tau^1, \tau^2)$ returns the number of positions at which two strings $\tau^1$ and $\tau^2$ differ; while SOFT_REGULAR$^{edit}(S)$ returns $\min\{E(\tau_\ell, \tau) \mid \tau \in L(M)\}$, where $E(\tau^1, \tau^2)$ returns the minimum number of insertions, deletions and substitutions to transform $\tau^1$ to $\tau^2$.*

**Theorem 18** *The cost functions SOFT_REGULAR$^{var}(S)$ and SOFT_REGULAR$^{edit}(S)$ are flow-based projection-safe.*

**Proof:** Van Hoeve et al. (2006) show that conditions 1 and 2 are satisfied. For example, consider the automaton $M$ shown in Figure 6 and $S = \{x_1, x_2, x_3\}$ with $D(x_1) = \{a\}$ and $D(x_2) = D(x_3) = \{a, b\}$. The flow networks corresponding to the SOFT_REGULAR$^{var}(S)$ and SOFT_REGULAR$^{edit}(S)$ functions are shown in Figure 7(a) and 7(b) respectively. The solid edges have zero weight and the dotted edges have unit weight. The thick edges show the flow corresponding to the tuple $(a, b, a)$.

The graphs are constructed as follows (van Hoeve et al., 2006): the vertices are separated into $n + 1$ layers, where $n = |\mathcal{X}|$, and each layer contains $|Q|$ nodes. The source $s$ is connected to $q_{0,0}$ at the first layer, and the sink $t$ is connected by $\{q_{n+1,i} \mid q_i \in F\}$ at the last layer. Between the $i^{th}$ and $(i + 1)^{th}$ layers, an zero weighted edge representing $v \in D(x_i)$ connects $q_{i,h}$ at the $i^{th}$ layer and $q_{i+1,k}$ at the $(i + 1)^{th}$ layer if $\delta(q_k, v) = q_h$. For SOFT_REGULAR$^{var}(S)$, a set of unit-weighted edges $E_{sub}$ is added to the graph, where $E_{sub} = \{(q_{i,k}, q_{i+1,h})_u \mid x_i \in \mathcal{X} \wedge u \in D(x_i) \wedge \exists v \neq u \ s.t. \ \delta(q_k, v) = q_h\}$. For SOFT_REGULAR$^{edit}(S)$, a set of unit-weighted edges $E_{edit}$ is added to the graph, where $E_{edit} = E_{sub} \cup \{(q_{i,k}, q_{i,h}) \mid x_i \in \mathcal{X} \wedge \exists v \ s.t. \ \delta(q_k, v) = q_h\} \cup \{(q_{i,k}, q_{i,k})_u \mid x_i \in \mathcal{X} \wedge u \in D(x_i)\}$.

Moreover, each assignment $\{x_i \mapsto v\}$ maps to a set of edges $\bar{E}$ labelled as $v$ at the layer $x_i$ in the networks. For example, $\{x_1 \mapsto a\}$ maps to the edges labeled as $a$ at the layer $x_1$ shown in Fig. 7(a). Thus, the SOFT_REGULAR cost functions satisfy condition 3 and are flow-based projection-safe. $\square$

For the SOFT_REGULAR cost functions, instead of the general flow computation algorithms, the dynamic programming approach can be applied to compute the minimum cost (van Hoeve et al., 2006; Demassey, Pesant, & Rousseau, 2006).

## 5.2 Experimental Results

In this section, a series of experiments with different benchmarks is conducted to demonstrate the efficiency and practicality of different consistencies with different global cost functions. We implemented the strong $\varnothing$IC, GAC*, FDGAC* and weak EDGAC* enforcement algorithms for these global cost functions in ToulBar2 version 0.5[1]. We compare their performance using five benchmarks of different natures. In case of the reified COP models, the instances are solved using ILOG Solver 6.0.

All benchmarks are crisp in nature, and are softened as follows. For each variable $x_i$ introduced, a random unary cost from 0 to 9 is assigned to each value in $D(x_i)$. Soft variants of global constraints are implemented as proposed. The target of all benchmarks is to find the optimal value within 1 hour.

---

[1]. http://carlit.toulouse.inra.fr/cgi-bin/awki.cgi/ToolBarIntro

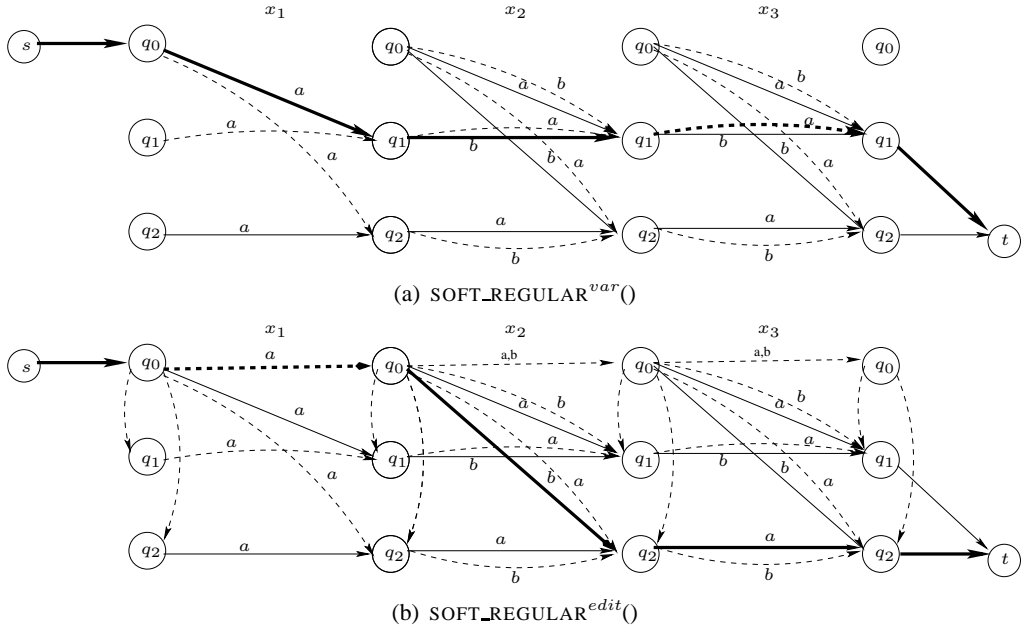

(a) SOFT_REGULAR$^{var}$()

(b) SOFT_REGULAR$^{edit}$()

Figure 7: The flow network corresponding to the soft REGULAR constraints

In the experiments, variables are assigned in lexicographical order. Value assignment starts with the value with minimum unary cost. The test was conducted on a Sun Blade 2500 ($2 \times 1.6$GHz USIIIi) machine with 2GB memory. The average runtime and number of nodes of five instances are measured for each value of $n$ with no initial upper bound. Entries are marked with a "*" if the average runtime exceeds the limit of 1 hour. The best results are marked using the '†' symbol.

### 5.2.1 BENCHMARKS BASED ON SOFT ALLDIFFERENT

The ALLDIFFERENT() constraint has various applications. In the following, we focus on two: the all-interval series and the Latin Square problem.

#### ALL INTERVAL SERIES

The all-interval series problem (prob007 in CSPLib) is modelled as a WCSP by two sets of variables $\{s_i\}$ and $\{d_i\}$ with domains $\{0, \ldots n-1\}$ to denote the elements and the adjacent difference respectively. Random unary costs ranging from 0 to 9 is placed on each variable. We apply two soft ALLDIFFERENT cost functions on $\{s_i\}$ and $\{d_i\}$ respectively, with a set of hard arithmetic constraints $d_i = |s_i - s_{i+1}|$ for each $i = 1, \ldots, n-1$.

The experiment is divided into two parts. We first compare results on enforcing different consistencies using global cost functions derived from ALLDIFFERENT(). Then we compare the result on using different approaches on modelling SOFT_ALLDIFFERENT$^{dec}$() functions.

The result of the first experiment is shown in Table 1, which agrees with the theoretical strength of the consistency notions as shown by the number of nodes. FDGAC* and GAC* always outperforms strong ∅IC and the reified modelling, but FDGAC* requires more time than GAC*. One explanation for this phenomenon is the problem structure. When $x_i$ and $x_{i+1}$ are assigned, $d_i$ is automatically assigned due to the hard constraint $d_i = |x_i - x_{i+1}|$. Thus, enforcing FDGAC* on the variables $\{d_i\}$ on every search node is not worthwhile.

(a) SOFT_ALLDIFFERENT$^{var}$()

| $n$ | Reified Approach | | Strong $\varnothing$IC | | GAC* | | FDGAC* | | Weak EDGAC* | |
|---|---|---|---|---|---|---|---|---|---|---|
| | Time(s) | Nodes | Time(s) | Nodes | Time(s) | Nodes | Time(s) | Nodes | Time(s) | Nodes |
| 8 | 1.3 | 571.0 | 0.2 | 296.4 | †0.1 | 181.0 | †0.1 | 86.4 | †0.1 | †15.4 |
| 9 | 3.9 | 1445.0 | 1.0 | 542.2 | 0.6 | 300.2 | 1.2 | 197.2 | †0.1 | †20.2 |
| 10 | 52.0 | 15860.6 | 20.2 | 5706.6 | 10.8 | 2589.4 | 15.2 | 1612.4 | †0.2 | †47.4 |
| 11 | 59.6 | 13286.2 | 31.8 | 7536.4 | 16.4 | 3273.6 | 21.0 | 1715.4 | †0.1 | †33.6 |
| 12 | 180.1 | 31015.2 | 77.8 | 12886.4 | 37.6 | 5204.6 | 46.8 | 2259.0 | †0.8 | †47.6 |

(b) SOFT_ALLDIFFERENT$^{dec}$()

| $n$ | Reified Approach | | Strong $\varnothing$IC | | GAC* | | FDGAC* | | Weak EDGAC* | |
|---|---|---|---|---|---|---|---|---|---|---|
| | Time(s) | Nodes | Time(s) | Nodes | Time(s) | Nodes | Time(s) | Nodes | Time(s) | Nodes |
| 8 | 1.6 | 777.0 | 0.2 | 396.8 | 0.2 | 219.6 | †0.1 | 93.8 | †0.1 | †16.0 |
| 9 | 3.9 | 1480.4 | 1.0 | 553.2 | 0.6 | 301.8 | 1.2 | 195.0 | †0.1 | †28.8 |
| 10 | 56.8 | 17753.8 | 21.2 | 5999.2 | 11.6 | 2654.6 | 16.0 | 1604.2 | †0.8 | †70.4 |
| 11 | 70.1 | 16149.6 | 38.4 | 9113.2 | 18.6 | 3551.8 | 23.0 | 1812.6 | †1.0 | †68.6 |
| 12 | 214.9 | 38438.6 | 96.4 | 16355.2 | 46.8 | 6405.0 | 52.6 | 2451.6 | †1.8 | †71.2 |

Table 1: The time (in seconds) and the number of nodes in solving the all-interval series instances

The second experiment is based on the following fact. The SOFT_ALLDIFFERENT$^{dec}(S)$ is flow-based projection-safe. It can be modelled as a flow network for consistency enforcement efficiently. Another way to model the global cost functions is to apply the decomposition directly. The cost returned by SOFT_ALLDIFFERENT$^{dec}(S)$ is equal to the sum of the costs returned by a set of soft binary cost functions $\{W_{i,j} \mid i > j \land x_i, x_j \in S\}$, where $W_{i,j}(a, b)$ returns 0 if $a \neq b$ and 1 otherwise. Thus, binary consistency notions, such as AC* and FDAC* can be applied directly.

We compare the performance on solving the all interval series problem with different modelling methods on SOFT_ALLDIFFERENT$^{dec}$(). The results are shown in Table 2. Under the same level of consistency, global cost functions remove an order of magnitude 10 to 100 times more nodes than the binary decomposition. However, the time required for binary cost functions is much smaller than global cost functions for AC* and FDAC*. This is because enforcing consistency notions on binary cost functions is faster than global cost functions, and the removal of nodes is not great enough to compensate the extra time for consistency enforcement of global cost functions. The runtime of weak EDGAC*, however, is the fastest among all (2 times over the EDAC* counterpart) since it is able to utilize global information to prune drastically more search space than any of the binary decomposition approaches.

## LATIN SQUARES

The Latin Square problem (prob003 in CSPLib) of order $n$ is to fill an initially empty $n \times n$ table using numbers from $\{0, \ldots, n - 1\}$ such that each number occurs once in every row and every column. We model and relax the problem as a WCSP by a set of variables $\{x_{ij}\}$ denoting the value placed in the cell at the $i^{th}$ row and the $j^{th}$ column with random unary costs. These costs are essentially restrictions/preferences on the value to be taken by each cell. Thus, our formulation can model different variants of the Latin Square problem, including the Latin Square Completion problem. One SOFT_ALLDIFFERENT() cost function is posted on the variables at each row and each column, denoting that same elements on the same rows and columns are allowed but with violation costs so that the resultant cost is optimal. The result is shown in Table 3, which is similar to Table 1. Besides, the runtime also agrees with the theoretical strength of the consistency notions.

| $n$ | Binary Decomposition | | | | | | Global Cost Functions | | | | | |
|---|---|---|---|---|---|---|---|---|---|---|---|---|
| | AC* | | FDAC* | | EDAC* | | GAC* | | FDGAC* | | Weak EDGAC* | |
| | Time(s) | Nodes | Time(s) | Nodes | Time(s) | Nodes | Time(s) | Nodes | Time(s) | Nodes | Time(s) | Nodes |
| 8 | †0.1 | 317.2 | †0.1 | 231.6 | †0.1 | 161.8 | 0.2 | 219.6 | †0.1 | 93.8 | †0.1 | †16.0 |
| 9 | †0.1 | 596.0 | †0.1 | 358.2 | †0.1 | 333.0 | 0.6 | 301.8 | 1.2 | 195.0 | †0.1 | †28.8 |
| 10 | 1.4 | 9113.8 | 1.0 | 5957.4 | 1.0 | 5483.2 | 11.6 | 2654.6 | 16.0 | 1604.2 | †0.8 | †70.4 |
| 11 | 1.6 | 7672.2 | 1.2 | 4578.4 | 1.2 | 4318.6 | 18.6 | 3551.8 | 23.0 | 1812.6 | †1.0 | †68.6 |
| 12 | 4.6 | 15897.2 | 3.2 | 10534.8 | 2.6 | 7414.4 | 46.8 | 6405.0 | 52.6 | 2451.6 | †1.8 | †71.2 |

Table 2: The time (in seconds) and the number of nodes in solving the all-interval series instances with different modelling

(a) SOFT_ALLDIFFERENT$^{var}$()

| $n$ | Reified Approach | | Strong ∅IC | | GAC* | | FDGAC* | | Weak EDGAC* | |
|---|---|---|---|---|---|---|---|---|---|---|
| | Time(s) | Nodes | Time(s) | Nodes | Time(s) | Nodes | Time(s) | Nodes | Time(s) | Nodes |
| 4 | 69.0 | 129958.0 | 1.8 | 3511.0 | †0.1 | 188.0 | †0.1 | 21.8 | †0.1 | †16.6 |
| 5 | * | * | 490.2 | 348790.4 | 26.0 | 12368.0 | †0.1 | 66.2 | †0.1 | †41.2 |
| 6 | * | * | * | * | * | * | 3.4 | 244.4 | †1.4 | †93.6 |
| 7 | * | * | * | * | * | * | 43.2 | 1429.4 | †16.2 | †425.2 |
| 8 | * | * | * | * | * | * | * | * | †148.2 | †2066.5 |

(b) SOFT_ALLDIFFERENT$^{dec}$()

| $n$ | Reified Approach | | Strong ∅IC | | GAC* | | FDGAC* | | Weak EDGAC* | |
|---|---|---|---|---|---|---|---|---|---|---|
| | Time(s) | Nodes | Time(s) | Nodes | Time(s) | Nodes | Time(s) | Nodes | Time(s) | Nodes |
| 4 | 62.7 | 121319.0 | 2.6 | 3859.8 | †0.1 | 187.6 | †0.1 | 21.8 | †0.1 | †16.6 |
| 5 | * | * | 531.4 | 376526.2 | 25.2 | 12254.0 | †0.1 | 66.2 | †0.1 | †41.2 |
| 6 | * | * | * | * | * | * | 3.4 | 244.4 | †1.4 | †93.6 |
| 7 | * | * | * | * | * | * | 43.4 | 1429.6 | †15.8 | †425.2 |
| 8 | * | * | * | * | * | * | * | * | †147.2 | †2066.5 |

Table 3: The time (in seconds) and the number of nodes in solving the Latin Square instances using SOFT_ALLDIFFERENT cost functions

The SOFT_ALLDIFFERENT$^{dec}$() cost functions can also be decomposed into binary disequality cost functions. We also perform experiments to compare the binary decomposition approach and our global cost function approach. The result is shown in Table 4. The result confirms that enforcing stronger consistency on global cost functions is efficient in terms of the number of nodes explored and also as the problem size grows large.

### 5.2.2 BENCHMARKS BASED ON SOFT GCC

The GCC() constraint has various applications. In the following, we focus on the Latin Square problem and round robin tournament problem.

| $n$ | Binary Decomposition | | | | | | Global Constraint Approaches | | | | | |
|---|---|---|---|---|---|---|---|---|---|---|---|---|
| | AC* | | FDAC* | | EDAC* | | GAC* | | FDGAC* | | Weak EDGAC* | |
| | Time(s) | Nodes | Time(s) | Nodes | Time(s) | Nodes | Time(s) | Nodes | Time(s) | Nodes | Time(s) | Nodes |
| 4 | †0.1 | 264.0 | †0.1 | 71.8 | †0.1 | 39.4 | †0.1 | 187.6 | †0.1 | 21.8 | 0.1 | †16.6 |
| 5 | 3.0 | 17955.8 | 0.4 | 3059.6 | †0.1 | 828.2 | 25.2 | 12254.0 | †0.1 | 66.2 | 0.1 | †41.2 |
| 6 | 639.2 | 2188035.4 | 167.8 | 346797.6 | 28.2 | 45817.8 | * | * | 3.4 | 244.4 | 1.4 | †93.6 |
| 7 | * | * | * | * | * | * | * | * | 43.4 | 1429.6 | †15.8 | †425.2 |
| 8 | * | * | * | * | * | * | * | * | * | * | †147.2 | †2066.5 |

Table 4: The time (in seconds) and the number of nodes in solving the Latin Square instances with different modelling

(a) SOFT_GCC$^{var}$

| $n$ | Reified Approach | | Strong ∅IC | | GAC* | | FDGAC* | | Weak EDGAC* | |
|---|---|---|---|---|---|---|---|---|---|---|
| | Time(s) | Nodes | Time(s) | Nodes | Time(s) | Nodes | Time(s) | Nodes | Time(s) | Nodes |
| 4 | 3.8 | 4865.6 | 2.8 | 3859.8 | †0.1 | 220.8 | †0.1 | 22 | †0.1 | †17.0 |
| 5 | 653.7 | 460989.2 | 621.2 | 376526.2 | 38.6 | 14482.8 | †0.1 | 66.2 | †0.1 | †48.2 |
| 6 | * | * | * | * | * | * | 4.8 | 244.6 | †1.2 | †87.0 |
| 7 | * | * | * | * | * | * | 58.4 | 1431.2 | †16.4 | †331.8 |
| 8 | * | * | * | * | * | * | * | * | †459.6 | †4730.8 |

(b) SOFT_GCC$^{val}$

| $n$ | Reified Approach | | Strong ∅IC | | GAC* | | FDGAC* | | Weak EDGAC* | |
|---|---|---|---|---|---|---|---|---|---|---|
| | Time(s) | Nodes | Time(s) | Nodes | Time(s) | Nodes | Time(s) | Nodes | Time(s) | Nodes |
| 4 | 2.2 | 2815.8 | 1.4 | 2326.6 | †0.1 | 131.8 | †0.1 | 20.4 | †0.1 | †17.0 |
| 5 | 165.2 | 122840.0 | 153.4 | 102493.6 | 10.0 | 4818.2 | †0.1 | 61.2 | †0.1 | †45.2 |
| 6 | * | * | * | * | 1407.4 | 357529.8 | 3.6 | 211.0 | †1.0 | †82.2 |
| 7 | * | * | * | * | * | * | 40.4 | 1243.6 | †13.4 | †318.4 |
| 8 | * | * | * | * | * | * | * | * | †285.2 | †3700.4 |

Table 5: The time (in seconds) and the number of nodes in solving the Latin Square instances using soft GCC constraints

## Latin Squares

We first focus on the Latin Square problem, which is described in Section 5.2.1. We use the same soft version but we replace SOFT_ALLDIFFERENT by either SOFT_GCC$^{var}$() or SOFT_GCC$^{val}$() cost functions which measure the violation differently. The results are shown in Table 5, which shows a similar result as Table 3. Weak EDGAC* always performs the best in terms of time and reduction in search space.

## Round Robin Tournament

The round robin problem problem (prob026 in CSPLib) of order $n$ is to schedule a tournament of $n$ teams over $n-1$ weeks. Each week is divided into $n/2$ periods, and each period is divided into two slots. A tournament must satisfy the following three constraints: (1) every team plays at least once a week, (2) every team plays at most twice in the same period over the tournament, and (3) every team plays every other team. Van Hentenryck, Michel, Perron, and Régin (1999) give a CSP model only based on GCC constraints: a triple of variables $(s_{ij}, t_{ij}, m_{ij})$ represents the match played on the $i^{th}$ week at the $j^{th}$ period. The assignment $\{s_{ij} \mapsto a, t_{ij} \mapsto b, m_{ij} \mapsto ab\}$ represents team $a$ is played against the team $b$. Ternary constraints link $s_{ij}$, $t_{ij}$ and $m_{ij}$ together such that $s_{ij}$ takes the value $a$

(a) soft_GCC$^{var}$

| $(N, P, M)$ | Reified Approach | | Strong ⌀IC | | GAC* | | FDGAC* | | Weak EDGAC* | |
|---|---|---|---|---|---|---|---|---|---|---|
| | Time(s) | Nodes | Time(s) | Nodes | Time(s) | Nodes | Time(s) | Nodes | Time(s) | Nodes |
| (4,3,2) | 1.7 | 1119.2 | 0.6 | 827.4 | 0.4 | 470.2 | 0.2 | 142.2 | †0.1 | †33.4 |
| (5,4,2) | 4.5 | 2016.6 | 2.2 | 1242.0 | 1.8 | 836.2 | 0.6 | 171.6 | †0.1 | †44.6 |
| (6,5,3) | * | * | * | * | * | * | * | * | †583.4 | †6508.8 |
| (7,5,3) | * | * | * | * | * | * | * | * | †1283.4 | †7476.6 |

(b) soft_GCC$^{val}$

| $(N, P, M)$ | Reified Approach | | Strong ⌀IC | | GAC* | | FDGAC* | | Weak EDGAC* | |
|---|---|---|---|---|---|---|---|---|---|---|
| | Time(s) | Nodes | Time(s) | Nodes | Time(s) | Nodes | Time(s) | Nodes | Time(s) | Nodes |
| (4,3,2) | 1.5 | 1046.8 | 0.4 | 794.6 | 0.4 | 464.6 | 0.2 | 141.0 | †0.1 | †33.0 |
| (5,4,2) | 3.5 | 1821.4 | 0.6 | 171.0 | 1.4 | 824.6 | 0.6 | 171.0 | †0.1 | †42.8 |
| (6,5,3) | * | * | * | * | * | * | * | * | †438.2 | †6499.6 |
| (7,5,3) | * | * | * | * | * | * | * | * | †765.0 | †7413.6 |

Table 6: The time (in seconds) and the number of nodes in solving the round robin tournament problems using soft_GCC cost functions

and $t_{ij}$ takes the value $b$ iff $m_{ij}$ takes the value $ab$ or $ba$. The first and the second requirements are represented by the GCC constraints on $\{s_{ij}, t_{ij} \mid i = w\}$ for each $w^{th}$ week and $\{s_{ij}, t_{ij} \mid j = p\}$ for each $p^{th}$ period. The third requirement is represented by a GCC constraint on $\{m_{ij}\}$.

The problem can be generalized by three parameters $(N, P, M)$: scheduling a tournament of $N$ teams over $M$ weeks, with each week divided into $P$ periods. Besides placing random unary costs, we also replace the GCC constraints by the soft variants. We try different combinations of $N$, $P$, and $M$. The results are shown in Table 6, which agrees with the theoretical strength of each consistency. It also shows that although enforcing stronger consistency is more expensive, it helps to reduce search space more. Thus, stronger consistency helps to solve larger instances.

### 5.2.3 Benchmarks Based on Soft Same

The same() constraint can be used to model the following two problems: (1) fair scheduling, and (2) people-mission scheduling.

#### Fair Scheduling

The problem is suggested in the Global Constraint Catalog[2]. The goal is to schedule $n$ persons into $s$ shifts over $d$ days such that the schedule is fair, *i.e.* each person should be assigned to the same number of the $i^{th}$ shift. For example, the schedule in Figure 8(a) is not fair. The person $p_1$ is assigned to the AM shift two times but $p_2$ is assigned to the AM shift once only. Figure 8(b) shows a schedule that is fair to everyone: both $p_1$ and $p_2$ are assigned to the AM shift and Overnight shift once, and the PM shift twice.

We model and soften the problem by a set of variables $\{x_{ij}\}$, which denote the shift assigned to the $i^{th}$ person on the $j^{th}$ day with random unary costs. The soft_same$^{var}(\{x_{p_1j}\}, \{x_{p_2j}\})$ cost functions are placed between each pair of persons $p_1$ and $p_2$, allowing violation for the fairness of the schedule to obtain minimum cost. We fix $s = 4$ and $d = 5$ and vary $n$. The results are shown in Table 7. Similarly to Table 5, weak EDGAC* produces the smallest number of nodes. However,

---

| | Day 1 | Day 2 | Day 3 | Day 4 |
|---|---|---|---|---|
| $p_1$ | AM | PM | PM | AM |
| $p_2$ | AM | PM | Overnight | PM |

(a) Unfair Schedule

| | Day 1 | Day 2 | Day 3 | Day 4 |
|---|---|---|---|---|
| $p_1$ | AM | PM | PM | Overnight |
| $p_2$ | AM | PM | Overnight | PM |

(b) Fair Schedule

Figure 8: Examples of Fair Scheduling

| $n$ | Reified Approach | | Strong $\varnothing$IC | | GAC* | | FDGAC* | | Weak EDGAC* | |
|---|---|---|---|---|---|---|---|---|---|---|
| | Time(s) | Nodes | Time(s) | Nodes | Time(s) | Nodes | Time(s) | Nodes | Time(s) | Nodes |
| 5 | 1983.9 | 1457812.6 | 74.2 | 20610.4 | 16.6 | 3511.8 | †0.1 | 27.4 | †0.1 | †25.4 |
| 6 | * | * | 1884.0 | 1038613.2 | 78.8 | 11031.8 | †0.4 | 40.4 | 1.0 | †34.0 |
| 7 | * | * | * | * | 377.0 | 36063.0 | †1.0 | 45.0 | 1.2 | †40.6 |
| 8 | * | * | * | * | 1630.0 | 124920.8 | †2.0 | 45.4 | 2.2 | †45.0 |
| 9 | * | * | * | * | * | * | †2.6 | †49.0 | 3.2 | †49.0 |
| 10 | * | * | * | * | * | * | †4.0 | 58.0 | 4.6 | †56.8 |
| 11 | * | * | * | * | * | * | †5.8 | 67.2 | 6.4 | †61.6 |

Table 7: The time (in seconds) and the number of nodes in solving the fair scheduling problem by enforcing different consistency notions.

weak EDGAC* requires more time to solve than FDGAC*. We look into the execution and discover that FDGAC* is so strong that the first lower bound computed is already very close, if not identical, to the objective value of the optimal solution. Therefore, enforcing weak EDGAC* gives only little improvement on reducing the search space.

People-Mission Scheduling

This problem extends the doctor-nurse rostering problem described by Beldiceanu, Katriel and Thiel (2004). Given three groups of $n$ persons, $m$ missions must be assigned to a team containing exactly one person in each group. We are also given a set of constraints restricting the combination of each team in one mission. The problem is to schedule those people into teams for missions such that no restriction is violated. We model the problem by $\{x_{ij}\}$ denoting the mission assigned to the $i^{th}$ person in the $j^{th}$ group with random unary costs. The combination restriction is softened as ternary cost functions. Two global cost functions SOFT_SAME$^{var}(\{x_{i1}\}, \{x_{i2}\})$ and SOFT_SAME$^{var}(\{x_{i2}\}, \{x_{i3}\})$ are posted to ensure each team exactly contains one person from each group. We fix $m = 6$ and vary $n$. The results are shown in Table 8. Similarly to Table 7, weak EDGAC* produces the smallest number of nodes, but requires more time than FDGAC*.

5.2.4 Benchmarks Based on Soft Regular

The REGULAR() constraint has many applications. In the following, we focus on two: (1) the nurse rostering problem, and; (2) the STRETCH() constraint modelling.

Nurse Rostering Problem

The nurse rostering problem (Cheng, Lee, & Wu, 1997) is to schedule a group of $n$ nurses into four shifts, PM shift, AM shift, Overnight, and Day-Off, over a period with most requirements satisfied.

| $n$ | Reified Approach | | Strong $\varnothing$IC | | GAC* | | FDGAC* | | weak EDGAC* | |
|---|---|---|---|---|---|---|---|---|---|---|
| | Time(s) | Nodes | Time(s) | Nodes | Time(s) | Nodes | Time(s) | Nodes | Time(s) | Nodes |
| 4 | 17.5 | 16992.0 | 4.0 | 5931.6 | 1.6 | 1517.4 | †0.2 | 247.8 | 0.4 | †238.8 |
| 5 | 427.8 | 283950.2 | 45.2 | 51029.8 | 11.2 | 7073.8 | †3.4 | 831.2 | †3.4 | †693.4 |
| 6 | * | * | 666.6 | 553001.2 | 156.6 | 75481.6 | †55.6 | 11065.2 | 69.2 | †10957.8 |
| 7 | * | * | * | * | * | * | †1348.0 | 333937.6 | 1714.0 | †296019.2 |

Table 8: The time (in seconds) and the number of nodes in solving the people-mission scheduling problem by enforcing different consistency notions.

(a) SOFT_REGULAR$^{var}$()

| $n$ | Reified Approach | | Strong $\varnothing$IC | | GAC* | | FDGAC* | | weak EDGAC* | |
|---|---|---|---|---|---|---|---|---|---|---|
| | Time(s) | Nodes | Time(s) | Nodes | Time(s) | Nodes | Time(s) | Nodes | Time(s) | Nodes |
| 3 | 260.66 | 118562 | 152.6 | 91661.4 | 2.0 | 956.2 | †0.1 | 28.6 | †0.1 | †22.8 |
| 4 | * | * | * | * | 25.4 | 6983.4 | †0.1 | 32.6 | †0.1 | †28.0 |
| 5 | * | * | * | * | * | * | 4.0 | 379.0 | †3.6 | †273.6 |
| 6 | * | * | * | * | * | * | 63.4 | 4017.6 | †37.8 | †1927.2 |
| 7 | * | * | * | * | * | * | 207.6 | 12242.0 | †42.8 | †2167.6 |
| 8 | * | * | * | * | * | * | 821.2 | 44414.0 | †229.2 | †10437.0 |

(b) SOFT_REGULAR$^{edit}$()

| $n$ | Reified Approach | | Strong $\varnothing$IC | | GAC* | | FDGAC* | | weak EDGAC* | |
|---|---|---|---|---|---|---|---|---|---|---|
| | Time(s) | Nodes | Time(s) | Nodes | Time(s) | Nodes | Time(s) | Nodes | Time(s) | Nodes |
| 3 | 286.6 | 122542.4 | 178.4 | 91933.8 | 9.2 | 2850.4 | †5.6 | 841.4 | 6.2 | †803.2 |
| 4 | * | * | * | * | 126.2 | 27267.6 | †25.4 | 2568.8 | 27.6 | †2424.0 |
| 5 | * | * | * | * | * | * | †535.6 | 47091.2 | 546.8 | †40244.0 |

Table 9: The time (in seconds) and the number of nodes in solving the nurse scheduling problem by enforcing different consistency notions.

In the experiment, the nurses are scheduled over four days such that (1) each nurse must have at most three AM shifts, at least two PM shifts, at least one Overnight, and at least one day-off; (2) each AM shift must have two nurses, each PM shift and each Overnight must have one nurse, and; (3) AM-shifts are preferred to be packed together, and the same preference is also posted on Day-Offs. We model this problem by a set of variables $\{x_{ij}\}$ to denote the shift assigned to the $i^{th}$ nurse on the $j^{th}$ day with random unary costs. Restrictions (1) and (2) are modeled by SOFT_GCC$^{val}$ cost functions, and (3) is modeled by either SOFT_REGULAR$^{var}$ or SOFT_REGULAR$^{edit}$ cost functions. All restrictions are allowed to be violated. The results are shown in Table 9. When SOFT_REGULAR$^{edit}$() is used, FDGAC* wins in term of runtime. However, if SOFT_REGULAR$^{var}$() is used, weak EDGAC* again requires the least time and the least number of nodes to solve.

## Modelling the Stretch() Constraint

Another application of the REGULAR() constraint is to model constraints that describe patterns. One example is the STRETCH() constraint.

(a) SOFT_REGULAR$^{var}$()

| $n$ | Reified Approach | | Strong ∅IC | | GAC* | | FDGAC* | | weak EDGAC* | |
|---|---|---|---|---|---|---|---|---|---|---|
| | Time(s) | Nodes | Time(s) | Nodes | Time(s) | Nodes | Time(s) | Nodes | Time(s) | Nodes |
| 30 | 183.5 | 7346.2 | 68.2 | 5203.2 | 36.4 | 573.0 | †30.0 | 171.4 | 35.2 | †162.6 |
| 35 | 419.4 | 13845.2 | 162.2 | 10297.8 | 80.6 | 971.6 | †57.6 | 239.8 | 69.0 | †233.4 |
| 40 | 842.4 | 23485.0 | 335.6 | 18067.2 | 148.4 | 1423.2 | †92.2 | 328.6 | 108.2 | †316.0 |
| 45 | 2318.2 | 55976.0 | 900.4 | 42007.0 | 378.2 | 3042.0 | †240.6 | 651.8 | 246.4 | †570.6 |
| 50 | * | * | 1142.2 | 88616.8 | 165.8 | 10762.2 | 130.2 | 1660.6 | †118.2 | †1316.0 |
| 55 | * | * | 2231.4 | 146901.6 | 306.0 | 17130.0 | 208.0 | 2291.8 | †193.8 | †1856.8 |

(b) SOFT_REGULAR$^{edit}$()

| $n$ | Reified Approach | | Strong ∅IC | | GAC* | | FDGAC* | | weak EDGAC* | |
|---|---|---|---|---|---|---|---|---|---|---|
| | Time(s) | Nodes | Time(s) | Nodes | Time(s) | Nodes | Time(s) | Nodes | Time(s) | Nodes |
| 30 | 216.2 | 6038.6 | 83.2 | 3861.6 | 40.6 | 447.4 | †34.2 | 123.8 | 39.6 | †122.4 |
| 35 | 561.6 | 12487.6 | 204.2 | 7626.0 | 86.8 | 706.0 | †60.6 | 164.0 | 70.8 | †162.8 |
| 40 | 1128.1 | 20585.8 | 413.0 | 12789.6 | 165.8 | 1080.0 | †90.8 | 208.4 | 101.6 | †194.0 |
| 45 | * | * | 1151.8 | 30480.6 | 446.4 | 2346.2 | 239.6 | 371.0 | †207.8 | †299.6 |
| 50 | * | * | 2122.8 | 62225.2 | 348.6 | 9189.0 | 204.8 | 967.6 | †185.0 | †823.2 |
| 55 | * | * | * | * | 623.8 | 13496.8 | 264.2 | 972.8 | †234.6 | †777.6 |

Table 10: The time (in seconds) and the number of nodes in solving the sliding problem by enforcing different consistency notions.

**Definition 24** *(Pesant, 2001) Given a value $v$ and a tuple $\ell \in \mathcal{L}(S)$. A $v$-stretch is the maximal subsequence of identical values $v$ in $\ell$. The* STRETCH($S$, $ub$, $lb$) *constraint is satisfied by $\ell$ if the length of the $v$-stretch in $\ell$ is at most $ub_v$ and at least $lb_v$.*

For simplicity, we omit the case when the STRETCH() constraint is circular. However, it can be handled by variable duplication (Pesant, 2004).

The STRETCH() constraint can be described by an automaton and thus modelled using the REGULAR() constraint (Pesant, 2004). The SOFT_REGULAR$^{var}$() and SOFT_REGULAR$^{edit}$() cost functions can be directly applied to define two soft variants of the STRETCH() constraint, namely SOFT_STRETCH$^{var}$() and SOFT_STRETCH$^{edit}$(). They are flow-based projection-safe by inheriting the same property from SOFT_REGULAR$^{var}$() and SOFT_REGULAR$^{edit}$() respectively.

To demonstrate the idea, we conduct experiments using the following sliding problem. The sliding problem of order $n$ consists a set of variables $\{x_1, \ldots, x_n\}$ with domains $D(x_i) = \{a, b\}$ and random unary costs. Each subsequence $\{x_i, \ldots, x_{n-5+i}\}$, where $1 \leq i \leq 5$, is required to contain $a$-stretches of length 2 and $b$-stretches of length 2 or 3. This restriction can be enforced through STRETCH constraints. We allow violations by modeling the constraints using either the SOFT_REGULAR$^{var}$ or SOFT_REGULAR$^{edit}$ cost functions. The results are shown in Table 10. Weak EDGAC* needs more time than FDGAC* when the instances are small, but weak EDGAC* pays off for large instances. This experiment also shows that the STRETCH constraint, an important constraint for modeling patterns, can be efficiently propagated in the WCSP framework.

### 5.2.5 DISCUSSIONS

A control comparison should have been conducted to examine the efficiency of ToulBar2 on the global cost functions encoded explicitly as tables as well. This cannot be done in a meaningful manner since the tables will be prohibitively large. Consider a simple cost function on 10 variables, each with a domain size of 10. The table already requires storage in the order of $10^{10}$ integers or tens of gigabytes.

Based on our experiments, two conclusions can be made. First, the experiments show that the reified approach and strong $\varnothing$IC are too weak both in terms of search space pruning and runtime reduction as compared to GAC*, FDGAC*, and weak EDGAC*. Second, the stronger consistency notions, weak EDGAC*, FDGAC* and GAC*, are worthwhile although they are more expensive to enforce. As shown from the experiments, GAC* reduces the number of search nodes at least 3 times more than the reified approach and 1.5 times more than strong $\varnothing$IC. GAC* has runtime at least 4 times less than the reified approach and 1.5 times less strong $\varnothing$IC. Weak EDGAC* and FDGAC* can reduce the search space by a much greater extent. Such additional pruning can usually compensate for the extra effort. Although Table 7 and Table 8 have shown cases where weak EDGAC* results in slower runtime, FDGAC* only wins by a small margin. In general, weak EDGAC* is still worthwhile to enforce. Table 10 further confirms that a stronger consistency is more desirable as the problem becomes large.

## 6. Conclusion and Remarks

In this section, we summarize our contributions and shed light on possible future directions of research.

Our contributions are five-fold. First, we introduce strong $\varnothing$IC based on $\varnothing$IC (Zytnicki et al., 2009) and give an algorithm to enforce strong $\varnothing$IC. Besides, we prove that strong $\varnothing$IC is confluent. We also show that enforcing strong $\varnothing$IC on a WCSP is stronger than GAC in the reified approach. Second, we give an algorithm to enforce GAC* for a WCSP, but enforcement is exponential. For efficient enforcement, we introduce *flow-based projection-safety*, which preserves the basic structure of global cost functions. We give sufficient conditions for a global cost function to be flow-based projection-safe. We also show as a part of the proof how projection and extension can be done so that the flow property is preserved. Third, we generalize FDAC* (Larrosa & Schiex, 2003) to FDGAC* and give an enforcement algorithm. Again, flow-based projection-safety helps FDGAC* enforcement. Fourth, we attempt to generalize EDAC* using similar methods, but find it to be nontrivial. We discover and give an example of a limitation of EDAC*. When cost functions share more than one variable, oscillation similar to the one demonstrated in Full AC* (de Givry et al., 2005) will occur. To solve this problem, we introduce cost-providing partitions, which restrict the distribution of costs when enforcing EDAC*. Based on cost-providing partitions, we define weak EDGAC*, which can be enforced in polynomial time for flow-based projection-safe global cost functions. Last but not least, we show that soft versions of ALLDIFFERENT(), GCC(), SAME() and REGULAR() are flow-based projection-safe. We also prove the practicality of our framework with empirical results on various benchmarks involving these global cost functions. The empirical results agree with the theoretical strength of the consistencies in terms of search tree pruning. The results also show that stronger consistency notions like weak EDGAC* and FDGAC* are more worthwhile to enforce, especially when solving large problems.

Three directions of future work are possible. The first one is to investigate if other even stronger consistency notions, such as VAC (Cooper et al., 2010), can also benefit from projection-safety to make their enforcement practical for global cost functions. Second, the current sufficient conditions for flow-based projection-safety might still be overly restrictive. For example, the global cost function SOFT_SEQUENCE (Maher, Narodytska, Quimper, & Walsh, 2008) does not satisfy the three conditions. It is interesting to find out other possible definition of flow-based projection-safety, which allow efficient projection and extension operations. Third, we only consider the minimum cost flow computation for finding the minimum cost in a global cost function. It is interesting to check if other approaches, such as mathematical programming, can be used to achieve the same results.

## Acknowledgments

Work described in this paper was generously supported by grants CUHK413808 and CUHK413710 from the Research Grants Council of Hong Kong SAR.

## References

Beldiceanu, N. (2000). Global Constraints as Graph Properties on a Structured Network of Elementary Constraints of the Same Type. In *Proceedings of CP'00*, pp. 52–67.

Beldiceanu, N., Carlsson, M., & Petit, T. (2004). Deriving Filtering Algorithms from Constraint Checkers. In *Proceedings of CP'04*, pp. 107–122.

Beldiceanu, N., Katriel, I., & Thiel, S. (2004). Filtering Algorithms for the Same Constraints. In *Proceedings of CPAIOR'04*, pp. 65–79.

Cheng, B., Lee, J. H. M., & Wu, J. (1997). A Nurse Rostering System Using Constraint Programming and Redundant Modeling. *IEEE Transactions on Information Technology in Biomedicine*, *1*, 44–54.

Cooper, M., de Givry, S., Sanchez, M., Schiex, T., Zytnicki, M., & Werner, T. (2010). Soft Arc Consistency Revisited. *Artificial Intelligence*, *174*, 449–478.

Cooper, M., & Schiex, T. (2004). Arc Consistency for Soft Constraints. *Artifical Intelligence*, *154*, 199–227.

Cooper, M. C. (2005). High-Order Consistency in Valued Constraint Satisfaction. *Constraints*, *10*(3), 283–305.

de Givry, S., Heras, F., Zytnicki, M., & Larrosa, J. (2005). Existential Arc Consistency: Getting Closer to Full Arc Consistency in Weighted CSPs. In *Proceedings of IJCAI'05*, pp. 84–89.

Demassey, S., Pesant, G., & Rousseau, L.-M. (2006). A Cost-Regular Based Hybrid Column Generation Approach. *Constraints*, *11*, 315–333.

Dijkstra, E. W. (1959). A Note on Two Problems in Connexion with Graphs. *Numerische Mathematik*, *1*, 269–271.

Johnson, D. (1977). Efficient Algorithms for Shortest Paths in Sparse Networks. *Journal of the ACM*, *24*(1), 1–13.

Larrosa, J., & Schiex, T. (2003). In the Quest of the Best Form of Local Consistency for Weighted CSP. In *Proceedings of IJCAI'03*, pp. 239–244.

Larrosa, J., & Schiex, T. (2004). Solving Weighted CSP by Maintaining Arc Consistency. *Artificial Intelligence*, *159*(1-2), 1–26.

Laurière, J.-L. (1978). A Language and a Program for Stating and Solving Combinatorial Problems. *Artificial Intelligence*, *10*, 29–127.

Lawler, E. (1976). *Combinatorial Optimization: Networks and Matroids*. Holt, Rinehart and Winston.

Leung, K. L. (2009). Soft Global Constraints in Constraint Optimization and Weighted Constraint Satisfaction. Master's thesis, The Chinese University of Hong Kong.

Maher, M., Narodytska, N., Quimper, C.-G., & Walsh, T. (2008). Flow-Based Propagators for the SEQUENCE and Related Global Constraints. In *Proceedings of CP'08*, pp. 159–174.

Pesant, G. (2001). A Filtering Algorithm for the Stretch Constraint. In *Proceedings of CP'01*, pp. 183–195.

Pesant, G. (2004). A Regular Language Membership Constraint for Finite Sequences of Variables. In *Proceedings of CP'04*, pp. 482–495.

Petit, T., Régin, J.-C., & Bessière, C. (2000). Meta-constraints on Violations for Over Constrained Problems. In *Proceedings of ICTAI'00*, pp. 358–365.

Petit, T., Régin, J.-C., & Bessière, C. (2001). Specific Filtering Algorithm for Over-Constrained Problems. In *Proceedings of CP'01*, pp. 451–463.

Régin, J.-C. (1996). Generalized Arc Consistency for Global Cardinality Constraints. In *Proceedings of AAAI'96*, pp. 209–215.

Régin, J.-C. (2002). Cost-Based Arc Consistency for Global Cardinality Constraints. *Constraints*, *7*, 387–405.

Sanchez, M., de Givry, S., & Schiex, T. (2008). Mendelian Error Detection in Complex Pedigrees using Weighted Constraint Satisfaction Techniques. *Constraints*, *13*(1), 130–154.

Schiex, T., Fargier, H., & Verfaillie, G. (1995). Valued Constraint Satisfaction Problems: Hard and Easy Problems. In *Proceedings of IJCAI'95*, pp. 631–637.

Van Hentenryck, P., Michel, L., Perron, L., & Régin, J.-C. (1999). Constraint Programming in OPL. In *Proceedings of the International Conference on the Principles and Practice of Declarative Programming*, pp. 98–116.

van Hoeve, W.-J., Pesant, G., & Rousseau, L.-M. (2006). On Global Warming: Flow-based Soft Global Constraints. *J. Heuristics*, *12*(4-5), 347–373.

Zytnicki, M., Gaspin, C., & Schiex, T. (2009). Bounds Arc Consistency for Weighted CSPs. *Journal of Artificial Intelligence Research*, *35*, 593–621.